\documentclass{ieeeaccess}
\usepackage{cite}
\usepackage{amsmath,amssymb,amsfonts}

\usepackage{algorithmic}
\usepackage{algorithm}
\usepackage{graphicx}
\usepackage{textcomp}

\usepackage{microtype}
\usepackage{subcaption}
\usepackage{multirow}
\usepackage{kotex}
\usepackage{tempora}
\usepackage{booktabs}
\usepackage{bm}
\usepackage{tabularx}
\usepackage{amsmath} 
\usepackage{amsfonts}      
\usepackage{soul}
\usepackage{url}

\newcommand\yj[1]{\textcolor{black}{#1}}
\newcommand\yjr[1]{\textcolor{black}{#1}}

\newcommand\jh[1]{\textcolor{black}{#1}}

\newcommand\jhthree[1]{\textcolor{black}{#1}}

\newcommand\cha[1]{{\textcolor{black}{#1}}}
\newcommand\chaaaaai[1]{{\textcolor{black}{#1}}}
\newcommand\jhnips[1]{\textcolor{black}{#1}}

\newcommand\chaaccess[1]{{\textcolor{black}{#1}}}
\newcommand\jhaaai[1]{\textcolor{black}{#1}} 
\newcommand\jhaccess[1]{\textcolor{black}{#1}} 

\renewcommand{\algorithmiccomment}[1]{\bgroup\hfill\scriptsize{//}~#1\egroup} 
\algsetup{linenosize=\scriptsize}

\newcolumntype{Y}{>{\centering\arraybackslash}X}
\newcolumntype{Z}{>{\centering\arraybackslash}p{0.15\textwidth}}

\usepackage[dvipsnames,table,xcdraw]{xcolor}

\def\BibTeX{{\rm B\kern-.05em{\sc i\kern-.025em b}\kern-.08em
    T\kern-.1667em\lower.7ex\hbox{E}\kern-.125emX}}
\begin{document}
\history{Date of publication 6, February, 2023, date of current version 23, February, 2023.}
\doi{10.1109/ACCESS.2023.3243108}

\def\etal{{\textit{et al.~}}}

\title{Observations on K-image Expansion of Image-Mixing Augmentation}
\author{
\uppercase{Joonhyun Jeong*}\authorrefmark{1},
\uppercase{Sungmin Cha*}\authorrefmark{2},
\uppercase{Jongwon Choi}\authorrefmark{3},
\uppercase{Sangdoo Yun}\authorrefmark{4},
\uppercase{Taesup Moon}\authorrefmark{2},
and \uppercase{Youngjoon Yoo$^{\dagger}$}\authorrefmark{1},
\IEEEmembership{Member, IEEE}}
\address[1]{ImageVision, NAVER Clova, Seongnam 13561,South Korea (e-mail: \{joonhyun.jeong, youngjoon.yoo\}@navercorp.com)}
\address[2]{Department of Electrical and Computer Engineering, Seoul National University, Seoul 08826, South Korea (e-mail: \{sungmin.cha, tsmoon\}@snu.ac.kr)}
\address[3]{Department of Advanced Imaging (GSAIM) and Graduate School of AI, Chung-Ang University, Seoul 06973, South Korea (e-mail: choijw@cau.ac.kr)}
\address[4]{AILab, NAVER Clova, Seongnam 13561, South Korea (e-mail: sangdoo.yun@navercorp.com)}

\markboth
{Author \headeretal: Preparation of Papers for IEEE TRANSACTIONS and JOURNALS}
{Author \headeretal: Preparation of Papers for IEEE TRANSACTIONS and JOURNALS}

\corresp{Corresponding author: YoungJoon Yoo (e-mail: youngjoon.yoo@navercorp.com). * denotes equal contribution.}

\begin{abstract}
Image-mixing augmentations (e.g., Mixup \yjr{and} CutMix), which typically \yjr{involve} mix\yjr{ing} two images, have become \yj{the} de-facto training \yjr{techniques} for image classification. Despite their huge success \yjr{in} image classification, the number of images to \yj{be} mix\yj{ed} has not been \yjr{elucidated in the literature:} only the naive K-image expansion \yjr{has been shown} to \yjr{lead to} performance degradation. This \yjr{study} derives a new K-image mixing augmentation based on the stick-breaking process under Dirichlet prior \yjr{distribution}. We demonstrate the superiority of our K-image expansion augmentation over conventional two-image mixing augmentation methods through extensive experiments and analyses: (1) more robust and generalized classifiers; (2) a more desirable loss landscape shape; (3) better adversarial robustness.
\yjr{Moreover}, we show that our probabilistic model can measure the sample-wise uncertainty and boost the efficiency for network architecture search \yjr{by achieving a 7-fold reduction in the} search time.
Code will be available at \url{https://github.com/yjyoo3312/DCutMix-PyTorch.git}.
\end{abstract}

\begin{keywords}
Image Classification, Augmentation, Dirichlet process
\end{keywords}

\titlepgskip=-15pt

\maketitle

\section{Introduction}
\yjr{The advent of deep classification networks has emphasized the} importance of data augmentation~\cite{devries2017cutout,yun2019cutmix,zhang2017mixup}. \yjr{P}roper data augmentation \yjr{can remedy} the performance degradation \yjr{due to} insufficient data and weak robustness to noisy data~\cite{chun2020empirical}. 
\yjr{Accordingly}, many \yjr{researchers} have proposed training strategies to \yjr{apply} the data augmentation methods \yjr{to the} deep classification network.

Among the \yjr{popular} data augmentation methods, image-mixing augmentation methods, especially  CutMix~\cite{yun2019cutmix} \yjr{exhibited} impressive performance \yjr{in} train\yjr{ing} large-scale deep classification networks. Image-mixing augmentation methods augment a new image by mixing the two paired images. 
For example, CutMix mixes the paired images by re-formulating their segments into one image. 
\yjr{By applying this} simple \yjr{princple}, the image-mixing augmentation successfully improves the performance of deep classification networks in various scenarios. 
Furthermore, \yjr{through} image-mixing augmentation, the deep learning model \yjr{becomes} robust to corrupted and uncertain data.

However, \yjr{the mechanism underlying} image-mixing augmentation is still not fully \yjr{understood}. Specifically, even the optimal number of images to mix has not been \yjr{elucidated:} \yjr{in response, the number of $K$ images was empirically set to 2}.
\yjr{Researchers}~\cite{(CoMixup)kim2021comixup,zhang2017mixup} \yjr{have made} \textit{naive} attempts \yjr{at} $K$-image expansions of the augmentation. 
\yjr{However}, the $K$-image expansion attempts ha\yjr{ve} been unsuccessful \yjr{in terms of} classification performance improvements.
Here, we aim to answer the following question: \yjr{Is} $K=2$, \yjr{the} number of image, optimal for image\jhthree{-mixing} augmentation?

\yjr{In this study, w}e derive a novel formulation \yjr{for} generaliz\yjr{ing} image-mixing augmentation \yjr{and apply it to} obtain improved results \yjr{for} image-mixing augmentation methods.
\yjr{Notably}, we find that a mixture of three or more images can further improve the performance of baseline methods using only the paired images. The superiority of the generalized formulation is validated \yjr{under different} classification scenarios. In addition, we \yjr{test the} robustness of our method\yjr{: the results reveal} that our method can drive the model into the widest (flattest) and deepest local minima.
\cha{In terms of adversarial robustness, we experimentally \yjr{demonstrate} \yj{that the proposed}  image-mixing augmentation methods \jhthree{strengthen} adversarial robustness and \yjr{reveal} that the expansion into the K-image case further improves the robustness.}

\yjr{Additionally}, we \yjr{demonstrate} that the proposed image-mixing augmentation can \yjr{be used to characterize} and estimate the uncertainty of the \jhthree{data} samples. Based on the estimated uncertainty, we acquire the subsampled data pool that can efficiently represent the overall data distribution. We \yjr{validate} the efficiency of our subsampling framework \yjr{under} the proposed scheme on network architecture search (NAS). \yjr{Notably, our} method preserves the performance \yjr{while achieving} 7.7 times \yjr{higher training} speed \yjr{when} using the subsampled data pool as a training set.

Our contribution can be summarized as follows. 
\begin{itemize}
  \item\yjr{We generalize} the image-mixing augmentations for image classification \yjr{and achieve} better generalization ability on unseen data against the baseline methods. 
  \item \cha{We experimentally analyze the \yjr{mechanism behind} the better generalization of K-image augmentation by illustrating a loss landscape \jhnips{near the discovered minima}. \yjr{Accordingly}, we \yjr{reveal} its \yjr{ability to achieve convergence to} \jhnips{wider} and deeper local minima. We also demonstrate that K-image augmentation improves the adversarial robustness of the model.}
  \item We propose a new data subsampling method by measuring sample uncertainty based on the proposed image-mixing augmentation, \yjr{which is }especially beneficial for handling a small number of training samples. We \jhthree{further} verify the efficiency of the proposed subsampling method by applying \yjr{it} to NAS.
\end{itemize}

\yjr{
The rest of the paper is organized as follows. In Section ~\ref{sec:related_work}, we list related studies of image augmentation, data efficiency, and architecture search. Section~\ref{sec:proposed_method} describes details of the formulation, implementation, and applications of the proposed augmentation method. Section~\ref{sec:experiment} demonstrates the experimental results of classification on CIFAR and ImageNet, adversarial robustness, and NAS from the subsampled dataset by the proposed methods.
Finally, we conclude the paper by mentioning the limitations in Section~\ref{sec:conclusion}.
}

\section{Related Works}
\label{sec:related_work}
\subsection{Augmentation}
\textcolor{black}{\yjr{Including} augmentation \yjr{in} training classification networks has become standard \yjr{practice for achieving} high performance.}
\yjr{Beginning with} simple augmentations, such as random crop, flipping, and color jittering, \yjr{increasingly complex techniques, including} including Cutout~\cite{devries2017cutout}, Mixup~\cite{zhang2017mixup}, CutMix~\cite{yun2019cutmix}, PuzzleMix~\cite{(PuzzleMix)kim2020puzzle}, SaliencyMix~\cite{(SaliencyMix)uddin2021saliencymix}, and Co-Mixup~\cite{(CoMixup)kim2021comixup} have been \yjr{applied}.
Among the \yjr{latter}, CutMix, Mixup, PuzzleMix, and SaliencyMix \yjr{typically} \textcolor{black}{mix} two images, and a recent variant, Co-Mixup, has reportedly \yjr{achieved an} impressive \yjr{enhancement in} classification performance.
\jhnips{
Co-Mixup also generalized the image-mixing augmentation methods into K-image cases using submodular-supermodular optimization, which \yjr{involves} huge computational cost. \yjr{Notably}, our proposed K-image mixing augmentation methods do not require optimization \yjr{and thus, have} less computational overhead \yj{than} Co-Mixup \yjr{while achieving similar performance} (Table \ref{table:compare_sota_mixing_augmentations})
}.
\subsection{Data Efficiency}
\yjr{S}everal approaches\yjr{, with the aim of efficiently utilizing a training dataset with a semantically important measure,} \textcolor{black}{have focused} on collecting examples that are considered informative by re-weighting the importance of \yjr{different} training samples: calculating the importance value from additional forward~\cite{not_all_samples_are_created_equal} and backward path~\cite{learning_to_reweight} of training, defining \yjr{the} approximated function~\cite{biasd_importance_sampling}, or using \textcolor{black}{loss} based training scheme~\cite{adaboost, hard_example_mining, focal_loss}. 
Nevertheless, a criterion based on the hardness of the example cannot be generalized \yjr{if the} samples \yjr{contain} label noise. \cite{self_pace_learning} also showed that hard examples\yjr{, unlike easy examples,} \yj{are unsuitable for the}  initial stages of training. \yjr{Building upon} previous works \yjr{on} measuring the importance of samples, we propose a robust importance subsampling methodology. 
We apply our subsampling concept to the differentiable search-based NAS \yjr{and achieve} performance improvements in both search time and classification accuracy.

\subsection{Network Architecture Search}
\yjr{Initiative NAS \cite{baker2016designing, zhong2018practical, zhong2020blockqnn, zoph2016neural, zoph2018learning} utilizing reinforcement learning (RL) requires significant computational cost so that is difficult to apply them to ImageNet scale dataset.}
 \yjr{To alleviate the problem, the weight-sharing NAS \cite{liu2018darts, xu2019pc, wu2019fbnet, xie2018snas, spos, rlnas,cai2018proxylessnas} introduce the \textit{SuperNet} concept, which incldues all the operation in the search space and extract the target architecture, \textit{SubNet} from the \textit{SuperNet}.}
 \yjr{For the extraction of the \textit{SubNet}, \cite{liu2018darts, xu2019pc, wu2019fbnet, cai2018proxylessnas} propose a gradient-based searching method, which has become dominant in the research field, currently.}
 \yjr{In this study, we demonstrate the effectiveness of the subsampled data from our proposed DCutMix in NAS by implementing it to PC-DART~\cite{xu2019pc}.  Like the other methods, PC-DARTS focuses on designing a cell, and a user can easily adapt the layer depth during the architecture search phase by appending or removing more of the search cells in the search space.}
\section{\chaaaaai{Proposed Method}}
\label{sec:proposed_method}
\chaaccess{In this section, we define a formulation of the proposed K-image mixing augmentation and apply a probabilistic augmentation framework to an image classification task. Accordingly, as a novel method of applying the proposed K-image mixing augmentation, we propose a subsampling method that utilizes the uncertainty measurement in the augmented data samples.}
\subsection{\chaaaaai{Formulation for K-image Mixing Augmentation}}
\noindent{\chaaaaai{\textbf{K-image mixing augmentation}}} \ \ 
\chaaccess{\yjr{In this subsection}, we formulate the K-image generalization for image mixing augmentation on the image classification task.}
We consider that augmented sample $x_c$ is composed of $x_1,..., x_K$, denoted as:
\begin{eqnarray}
\label{eq:composite}
x_c =  f_c(x_1,...,x_K; \phi_1,...,\phi_K),
\end{eqnarray}
where the function $f_c(\cdot)$ \yjr{denotes} the composite function, and the term  $\bm{\phi}=\{\phi_1,...,\phi_K\}$ is a mixing parameter denoting the portion of each sample $x_k$ on the composite sample $x_c$.
\chaaccess{Note that Equation (\ref{eq:composite}) can be considered as the general form of the popular image-mixing augmentations, such as \chaaaaai{ CutMix \cite{yun2019cutmix} and Mixup \cite{zhang2017mixup}} \yjr{which mix only two images.} 
Specifically, in the case of Mixup (denoted as \textit{DMixup})} the function $f_c(\cdot)$ is defined by the weighted summation as follows:
\begin{eqnarray}
\label{eq:mixup}
x_c =  \Sigma^{K}_{k=1}\phi_k x_k.
\end{eqnarray}
The mixing parameter $\bm{\phi}$ is defined by the Beta distribution in the usual two-image cases (\textit{e.g.}, CutMix~\cite{yun2019cutmix} and Mixup~\cite{zhang2017mixup})
\chaaccess{Note that Equation (\ref{eq:mixup}) can be naturally expanded to the case of the K-image mixing case by applying a Dirichlet distribution \jhnips{$\bm{\phi}\sim Dir(\bm{\alpha}), \bm{\alpha}\in\mathbb{R}^K$.}}
\chaaccess{In this case, the composite sample $x_c$ becomes a random variable for the given hyper-parameter $\bm{\alpha}$.}

\noindent{\chaaaaai{\textbf{K-image generalization of CutMix}}} 
\chaaccess{Based on the above formulation, we define the $K$-image generalization of Cutmix (denoted as \textit{DCutMix}).
Note that the definition of the function $f_c(\cdot)$ becomes} more complicated \jh{because} the function should contribute all the \jh{segments} of images $x_1,..x_K$ to composite image $x_c$ \yjr{considering} their mixing parameters $\bm{\phi}$.
\chaaccess{Here, we composite the images proportionally following the stick-breaking process (SBP~\cite{sethuraman1994constructive}), with the widely used approach of sampling from Dirichlet distribution.}

\begin{figure}[t!]
\centering
    \includegraphics[width=0.99\linewidth]{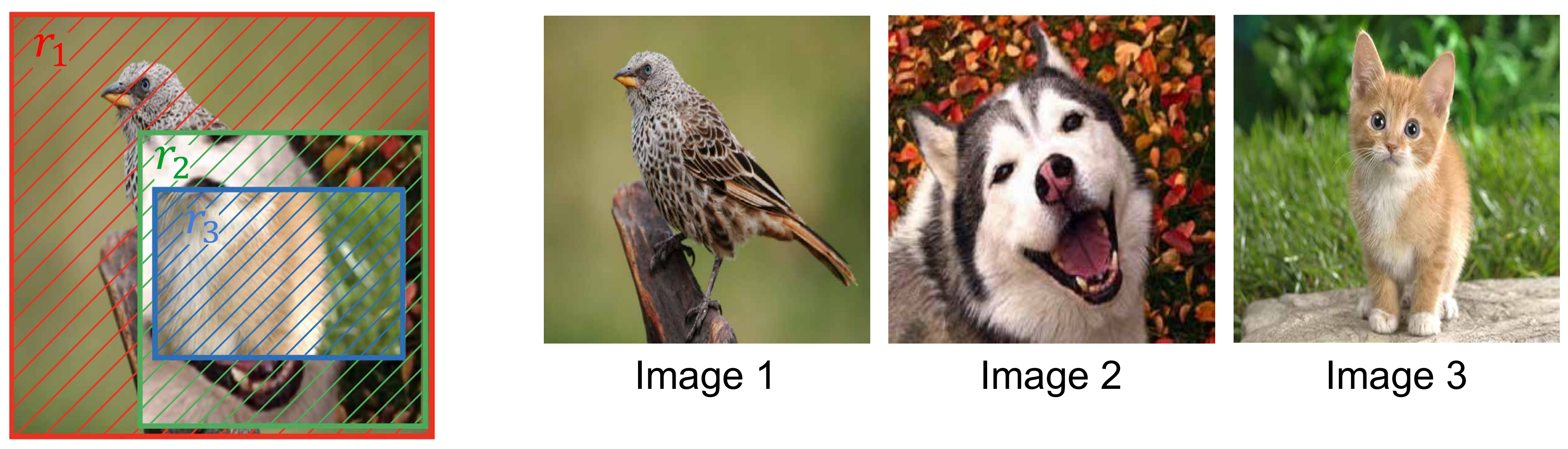}
    \centering
    \caption{Example of three-image composition in CutMix case. We composite red box to green box, green box to blue box with the ratio of $1:1-v_1$, and $1:1-v_2$ \jhnips{as in} (\ref{eq:cut_mix_sbp}). \jhnips{Consequently}, the region proportion of each image fraction $r_1$ \jhnips{(red diagonal pattern)}, $r_2$ \jhnips{(green diagonal pattern)}, and $r_3$ \jhnips{(blue diagonal pattern)} will correspond to $\{\phi_1, \phi_2, \phi_3\}$, \jhnips{which follows Dirichlet distribution.}
    Notably, a low variable anchor image (anchor image with red border) mostly serves as either an easy or hard sample regardless of the occlusion position. In contrast, a highly variable anchor image (anchor image with green border) serves as both an easy and hard sample depending on the random image mixing operation. These highly variable anchor images possibly provide more diverse information during training.}
    \label{fig:composition_toy_examples}
\end{figure}

\chaaccess{Assume that $\bm{\phi}$ is sampled from the prior distribution $Dir(\bm{\alpha})$.}
\chaaccess{The K-image mixing augmentation of CutMix is conducted by compositing the image with respect to the proportion $\phi_k\in\bm{\phi}$, where $\Sigma_k \phi_k=1$.}
\jhaccess{For sampling $\bm{\phi}$ from $Dir(\bm{\alpha})$, we use SBP by leveraging an intermediate variable $\bm{v}$ as follows.}
\chaaccess{Firstly, let $\bm{v}=[v_1,...,v_{K-1}]\in \mathbb{R}^{K-1}$ and each $v_k$ is denoted as follows:}
\begin{eqnarray}
\begin{aligned}
\label{eq:stick_breaking}
v_1 &= \phi_1\\
v_k &= \phi_k / \prod^{k-1}_{j=1} (1 - v_j), ~k=2,...,K-1.
\end{aligned}
\end{eqnarray}

\jhnips{Note that the variable $v$ is 
sampled from the beta distribution ($Beta(1, \alpha)$) by deriving SBP.}
Now, we define the image fractions $\bm{r} = \{r_1,...,r_K\}$ from \jhnips{K different images which constitute to a mixed sample $x \in \mathbb{R}^{W \times H \times C'}$}. 
Let the function $\Tilde{r} = d(x|v)$ randomly discriminate the \jhnips{image fractions $\Tilde{r}:x\backslash \Tilde{r}$} with the area ratio $v:1-v$, \chaaccess{where $x\backslash \Tilde{r}$ \yjr{denotes} the region of $x$ excluding $\Tilde{r}$.
Consequently, the fractions $\bm{r}$ are determined by following equation:}
\begin{eqnarray}
\label{eq:cut_mix_sbp}
r_k &= d(x\backslash \sum^{k-1}_{j=0}r_j|v_k), ~k=1,...,K-1,
\end{eqnarray}
 \chaaccess{where the virtual fraction $r_0$ and the last fraction $r_K$ are set to $\emptyset$ and $x\backslash\sum^{K-1}_{j=1}r_j$, respectively.} 
 \chaaccess{The discrimination function $d(\cdot)$ determines the exact bounding box coordinates $r_{k_x}, r_{k_y}, r_{k_w}, r_{k_h}$ of image fraction $r_k$, to be located within the bounding box coordinates of former image patch $r_{k-1}$. These coordinates are randomly sampled from the uniform distribution with random variable $\gamma$, as follows:}
\begin{eqnarray}
\begin{aligned}
\label{eq:dcutmix_gamma}
r_{k_x} \sim Unif(r_{k-1_x}, r_{k-1_x} + r_{k-1_w} - r_{k_w}),\\
r_{k_y} \sim Unif(r_{k-1_y}, r_{k-1_y} + r_{k-1_h} - r_{k_h})
\end{aligned}
\end{eqnarray}
 \chaaccess{where its width $r_{k_w}$ and height $r_{k_h}$ are determined by $v_k$ as defined in Equation (\ref{eq:cut_mix_sbp}).
Note that, in the case of $k=1$, $r_{1_x} = 0$, $r_{1_y} = 0$, $r_{1_w} = W$ and $r_{1_h} = H$.
Hence, the composite function $f_c$ of the \jh{DCutMix} is governed by \jh{hyper-parameter $\bm{\alpha}$ and random variable $\gamma$.}
An illustration of the proposed K-image mixing augmentation following Equation Equation (\ref{eq:cut_mix_sbp}) is presented in Figure~\ref{fig:composition_toy_examples}.}

\chaaccess{In the subsequent experiment section, we will experimentally demonstrate the advantages of the proposed K-image generalization in terms of loss landscape, adversarial robustness, and classification accuracy.}

\noindent\textbf{Probabilistic framework}
\chaaccess{The overall probabilistic framework of the classification problem, considering the proposed augmentation, can be defined as: }
\begin{eqnarray}
\begin{aligned}
\label{eq:lik_mixup}
p(l_c|x_c) &= \int p(l_c|x_c, \bm{\phi})p(\bm{\phi}|\bm{\alpha})d\bm{\phi},\\
&= \int p(\Sigma_k\{\phi_k l_k\}|f_c(\textcolor{black}{\bm{x}};\bm{\phi}))p(\bm{\phi}|\bm{\alpha})d\bm{\phi},\\
&\cong \frac{1}{N_s}\sum_{ \textcolor{black}{ \bm{\phi}^{(j)}} }f_W(\{\Sigma_k{\phi^{(j)}_k l_k}\}|f_c(\textcolor{black}{\bm{x}};\bm{\phi}^{(j)})),
\end{aligned}
\end{eqnarray}
where $(\bm{x},\bm{l})=\{(x_1,l_1),...,(x_K, l_K)\}$, $l_c=\Sigma_k\{\phi_k l_k\}$, and \textcolor{black}{$\bm{\phi}^{(j)}$} is the $j^{th}$ sample drawn from the Dirichlet prior distribution $p(\cdot|\bm{\alpha})$.
Hereafter, we define the label \jhnips{$l_i\in\mathbb{R}^L$} as a one-hot indexing variable denoting one of total $L$ \yjr{total} classes.
\chaaccess{Based on the derivation from Monte-Carlo dropout~\cite{gal2016dropout}, we can \jh{approximate} the distribution $p(l_c|x_c)$ in  the variational function $f_W(\cdot)$ \jh{with regard to several different $\bm{\phi}$ and $\gamma$ samples}. The variational function is realized by a classification network, parameterized by $W$, with a softmax output.}
\chaaccess{In the case of DCutMix, we additionally consider another variable $\gamma$ from  (\ref{eq:lik_mixup}), such as:}

\begin{eqnarray}
\begin{aligned}
\label{eq:lik_CutMix}
p(l_c|x_c) \cong \sum_{\gamma^{(i)}}\sum_{\bm{\phi}^{(j)}}f_W(\{\Sigma_k{\phi^{(j)}_k l_k}\}|f_c(\textcolor{black}{\bm{x}};\bm{\phi}^{(j)}, \gamma^{(i)})).
\end{aligned}
\end{eqnarray}

\chaaccess{Consequently, from (\ref{eq:lik_mixup}) and (\ref{eq:lik_CutMix}), we can approximate the posterior $p(l_c | x_c)$ by estimating the predictive mean of network outputs, depending on several differently augmented data sampled \yjr{for varying} $\bm{\phi}$ and $\gamma$ \yjr{values}. Similarly, the uncertainty of a given data sample $x_c$ for the given classification network can be approximated by calculating the posterior estimated from augmented data samples.}

\jhnips{\subsection{Implementation Details of DCutMix}}
\chaaccess{In this section, we present the implementation details of DCutMix. We describe the pseudo-code of the mixing process of DCutMix in Algorithm \ref{alg:dcutmix}. First, we sample variable $\phi$ from $Dir(\bm{\alpha})$ (see Line \jhaccess{1}). For $K-1$ iterations, we cut and mix $K-1$ image fractions. At each iteration, a mini-batch input and target are shuffled along with the batch dimension. An intermediate variable $v$ is then selected using SBP sampled from $\phi$ (see Line 6, 7, 12, and 13). \jhaaai{The} variable $v$ determines the width and height of the \jhaaai{image patch to be mixed}, where the exact position is bounded on the former image patch (see Lines 17 and 18). We then cut an image patch from source images $x_s$ and mix on $x_c$ (see Line 19). In Lines 20-27, the soft label is accordingly mixed by $\lambda$ and $\lambda_{K-1}$, which denote the exact area ratio of each mixed image patch.}

\begin{figure}[t!]
\centering
    \includegraphics[width=0.99\linewidth]{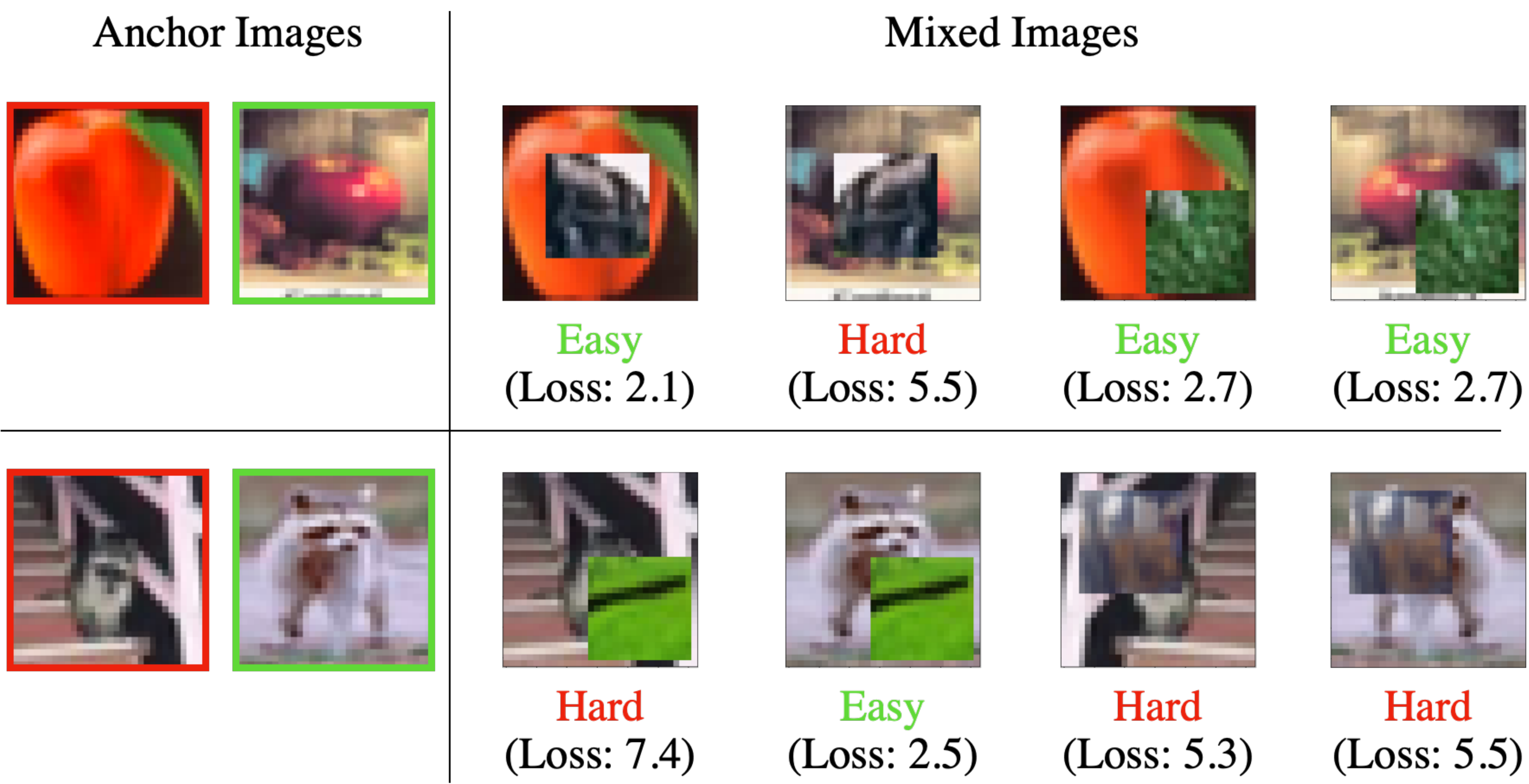}
    \centering
    \caption{Example of composited samples where \jhnips{the loss values are highly variable or invariable} depending on the position of occlusion by non-anchor image patches.
     The anchor images (bordered with green) have highly variable loss values depending on the occlusion. In contrast, the other anchor images (bordered with red) \jhnips{have relatively invariable loss values regardless of the occlusion.}}
    \label{fig:sampling_toy_examples}
\end{figure}

\begin{algorithm}[h]
\caption{Pseudo-code of DCutMix}
\label{alg:dcutmix}
\begin{algorithmic}[1] 
\REQUIRE Input images $x \in \mathbb{R}^{N \times C \times W \times H}$, Target labels $l \in \mathbb{R}^{N \times L}$, Number of mixing images $K$, Parameter of Dirichlet distribution $\bm\alpha \in \mathbb{R}^{N \times K}$\\
\ENSURE Augmented image $x_{c}$, Augmented label $l_{c}$
\STATE $\phi = Dir(\bm{\alpha})$
\STATE $x_c = x $
\FOR {$k \gets 1$ to $K-1$}
\STATE $x_s, l_s = $Shuffle$(x, l)$\\
\yjr{// stick-breaking process from equation~(\ref{eq:stick_breaking})}
\IF{$k == 1$}
    \STATE $v = \phi[k]$ \COMMENT{sample intermediate variable $v$.}
    \STATE $v_{m} = 1 - v$
    \STATE $r_{w,p}, \ r_{h,p} = W, \ H$ \COMMENT{set the size of bounding box.}
    \STATE $r_{x,p}, \ r_{y,p} = 0, \ 0$ 
    \STATE $l_p = l$  \COMMENT{set a label of the previous image patch.}
\ELSIF{$k \leq K-1$}
    \STATE $v = \phi[k] / v_m$
    \STATE $v_m = v_m \times (1 - v)$
\ENDIF
\\
\yjr{// bounding box setting, from equation~(\ref{eq:cut_mix_sbp}) and (\ref{eq:dcutmix_gamma}).}
\STATE $r_w = $Round$(r_{w,p} * \sqrt{1 - v})$
\STATE $r_h = $Round$(r_{h,p} * \sqrt{1 - v})$ 
\STATE $r_x\sim Unif(r_{x,p}, r_{x,p} + r_{w,p} - r_w)$
\STATE $r_y \sim Unif(r_{y,p}, r_{y,p} + r_{h,p} - r_h)$
\STATE $x_{c}[:, :, r_x:r_x+r_w, r_y:r_y+r_h] = x_s[:, :, r_x:r_x+r_w, r_y:r_y+r_h]$
\STATE $\lambda = (r_{w,p} \times r_{h,p} - r_w \times r_h) / W \times H$ \COMMENT{set the mixing ratio.}
\IF{$k == 1$} 
    \STATE $l_c = \lambda \times l_p$ \COMMENT{make a soft label based on the mixing ratio.}
\ELSIF{$ k < K-1$}
    \STATE $l_c = l_c + \lambda \times l_p$
\ELSE
    \STATE $\lambda_{K-1} = (r_w \times r_h) / (W \times H)$ 
    \STATE $l_c = l_c +  \lambda \times l_p + \lambda_{K-1} \times l_s$
\ENDIF
\STATE $r_{w,p}, \ r_{h,p}  = r_w, \ r_h$ \COMMENT{record the bounding box and soft label.}
\STATE $r_{x,p}, \ r_{y,p} = r_x, \ r_y$
\STATE $l_p = l_s$
\ENDFOR
\STATE \textbf{Return} $x_c$, $l_c$ \COMMENT{return an augmented image and soft label.}
\end{algorithmic}
\end{algorithm}

\subsection{\chaaaaai{Subsampling using the Measured Data Uncertainty}}
\chaaaaai{As a new \yjr{method of utilizing the}  K-image mixing augmentation, we propose a novel subsampling method that considers the data uncertainty \jhnips{obtained from K-image augmentation} for the first time.
In order to measure the uncertainty of a data sample, we define the loss distribution $L(l_c | x_c)$ for variously augmented data samples \jhnips{depending on $\bm{\phi}, \gamma$} and its expectation can be approximated based on (\ref{eq:lik_CutMix}) as follows:}
\jh{
\begin{eqnarray}
\begin{aligned}
\label{eq:exploss_CutMix}
\mathbb{E}[L(l_c|x_c)] \cong \sum_{\gamma^{(i)}}\sum_{\bm{\phi}^{(j)}}\mathcal{L}(\{\Sigma_k{\phi^{(j)}_k l_k}\}|f_c(\bm{x};\bm{\phi}^{(j)}, \gamma^{(i)})),
\end{aligned}
\end{eqnarray}
where $\bm{x}=\{x_1,..,x_K\}$, and $\mathcal{L}$ denotes the cross-entropy loss. The expectation is defined on the space by the random variable $\bm{\phi}$ and $\gamma$. Similarly, the uncertainty can also be acquired by estimating the variance of the loss distribution $L$. Figure~\ref{fig:sampling_toy_examples} shows qualitative examples of uncertainty measurement,  given sample data and their mixed images. \yjr{Noticeably,} the diverse tendency of loss values changes for each mixed image, mainly depending on the randomly selected position of occlusion caused by non-anchor image patches.
}

\jh{
For measuring the sample-wise uncertainty using the loss distribution, we select an \chaaaaai{\textit{anchor sample}}
$x_i\in \bm{x}$ with fixed $\phi_i$ and then jitter $\bm{\phi}\backslash\phi_i$ related to other non-anchor samples $\bm{x}\backslash x_i$ to calculate the uncertainty of the anchor sample $x_i$. 
The $\bm{\phi}\backslash\phi_i$ are drawn from a conditional Dirichlet distribution $D(\bm{\alpha}\backslash\alpha_i)$, \yjr{according to} its definition. We will term $L_i = \{\mathcal{L}_{i,m}|m=1,...M\}$ as the loss distribution for all the mixed images given the anchor $x_i$ \yjr{the corresponding loss} is  calculated from (\ref{eq:exploss_CutMix}). The number $M$ denotes the total number of sampling $\bm{\phi}\backslash\phi_i$ from $D(\bm{\alpha}\backslash\alpha_i)$.
}

Based on the sample-wise uncertainty measurement, we aim to sample the core \jhnips{training} data sub-set among the entire \jhnips{training} dataset. \jhnips{\yjr{Presumably,} for better generalization of a neural network when training using a small number of data points and image-mixing augmentation, the core training sub-set should consist of the highly uncertain samples which can serve as both easy- and hard-level samples depending on the image-mixing augmentation (e.g., the images bordered with green in Figure \ref{fig:sampling_toy_examples}).}
\jhnips{Therefore, a new training subset is subsampled by descending order of uncertainty measure. 
We observed that employing the coefficient of variation (CV) metric, which is defined as $\frac{\sigma(L_{i})}{m(L_{i})}$
where $\sigma(\cdot)$ and $m(\cdot)$ is the standard deviation and average of $L_{i}$, is most effective for measuring the uncertainty (See \jhnips{Figure \ref{fig:various_sampling_methods} for details})}.

\subsubsection{Subsampling details}
We \yjr{herein} describe the implementation details for the proposed subsampling framework. 
\jhnips{
A newly subsampled set $D$ for each class is defined as follows:
\begin{eqnarray}
\label{eq:dataset_distillation}
D = S(O(L_{k}) | k=j_{1}, ..., j_{N_{intra}}, t),
\end{eqnarray}
where $O(\cdot)$ denotes the subsampling measure and $S(\cdot)$ denotes a sampling function indicating whether data sample $x_{k}$ is to be included in $D$ or not by using the subsampling ratio $t$. Here, $j$ denotes the index of $N_{intra}$ number of intra-class images where the class labels are equivalent among the others. $O(\cdot)$ is a proxy for subsampling; data samples are subsampled \jhaaai{in} order of $O$. With regard to the sampling function $S(\cdot)$, it samples $t \times N_{intra}$ data samples based on the sampling measure $O(\cdot)$\jhnips{, which falls into two categories: a \textit{deterministic} function sampling top \textcolor{black}{$t \times N_{intra}$} samples \jhaaai{sorted} by $O(\cdot)$, and an \textit{interval-based} function that collect samples \jhaaai{sorted} by \jhaaai{$O(\cdot)$} with a fixed interval.}
}

\jhaaai{Regarding the subsampling measure $O(\cdot)$, we employed the sample-wise uncertainty measure using Coefficient Variation (CV), $\frac{\sigma(L_{i})}{m(L_{i})}$ (where $L_i$ is derived from (\ref{eq:exploss_CutMix}).} For estimating the sample-wise uncertainty, we set the number of non-anchor images $\bm{x} \backslash x_i$ and their Dirichlet sampling parameter $\bm{\alpha} \backslash \alpha_i$ as 2 and $\{\frac{2}{9}, \frac{2}{9}\}$, respectively. Additionally, we set the total number of sampling $\bm{\phi} \backslash \phi_i$ from the Dirichlet distribution, namely $M$, as 10. 

\begin{table}[t]
    \begin{minipage}{.5\textwidth}
        \small
        \centering
        \tabcolsep=0.07cm
                \begin{tabular}{@{}lccc@{}}
                \toprule
                \multirow{2}{*}{Model}            & \multirow{2}{*}{\# Params} & \multirow{2}{*}{\begin{tabular}[c]{@{}c@{}}Top-1 \\ Err (\%)\end{tabular}} &  \\
                                                  &                             &                           \\ \midrule
                PyramidNet-110 ($\Tilde{\alpha}=64$)~\cite{han2017deep}          & 1.7 M     & 19.85 \\
                + Mixup~\cite{zhang2017mixup}                                   & 1.7 M     & 18.92 (-0.93) \\
                + CutMix~\cite{yun2019cutmix}                                   & 1.7 M     & 17.97 (-1.88) \\
                + DMixup ($K=3, \alpha=\frac{1}{3}$)                                                         & 1.7 M     &  \textbf{18.60 (-1.25)} \\
                + DCutMix ($K=5, \alpha=0.2$)                                                      & 1.7 M     & \textbf{16.95 (-2.90)} \\
                \midrule
                PyramidNet-200 ($\Tilde{\alpha}=240$)                                   & 26.8 M     & 16.45 \\
                + StochDepth \cite{stochasticdepth}                                     & 26.8 M     & 15.86 (-0.59) \\
                + Label Smoothing \cite{szegedy2016rethinking}                          & 26.8 M     & 16.73 (+0.28) \\
                + Cutout \cite{devries2017cutout}                                       & 26.8 M     & 16.53 (+0.08) \\
                + DropBlock \cite{ghiasi2018dropblock}                                  & 26.8 M     & 15.73 (-0.72) \\
                + Mixup \cite{zhang2017mixup}                                           & 26.8 M     & 15.63 (-0.82) \\
                + Manifold Mixup \cite{verma2018manifoldmixup}                          & 26.8 M     & 15.09 (-1.36) \\
                + CutMix \cite{yun2019cutmix}                                           & 26.8 M     & 14.47 (-1.98) \\
                {+ StyleMix} \cite{hong2021stylemix}                                                             & {26.8 M}     & {16.37 (-0.08)} \\
                {+ StyleCutMix} \cite{hong2021stylemix}                                                               & {26.8 M}     & {14.17 (-2.28)} \\
                + DMixup ($K=3, \alpha=1$)                                                               & 26.8 M     & \textbf{15.07 (-1.38)} \\
                
                + DCutMix ($K=5, \alpha=1$)                                                              & 26.8 M     & \textbf{13.86 (-2.59)} \\
                \bottomrule
                \end{tabular}
    \end{minipage}
    \begin{minipage}{.5\textwidth}
        \small
        \centering
        \tabcolsep=0.05cm
                \begin{tabular}{@{}lccc@{}}
                \toprule
                \multirow{2}{*}{Model}            & \multirow{2}{*}{\begin{tabular}[c]{@{}c@{}}Top-1 \\ Err (\%)\end{tabular}} &  \\
                                                  &    \\ \midrule
                                                  
                PyramidNet-200 ($\Tilde{\alpha}=240$)                               & 3.85 \\
                PyramidNet-200 + Cutout \cite{devries2017cutout}                    & 3.10 (-0.75) \\
                PyramidNet-200 + Mixup \cite{zhang2017mixup}                        & 3.09 (-0.76) \\
                PyramidNet-200 + Manifold Mixup \cite{verma2018manifoldmixup}       & 3.15 (-0.7) \\
                PyramidNet-200 + CutMix \cite{yun2019cutmix}                        & 2.88 (-0.97) \\
                PyramidNet-200 + DMixup ($K=5, \alpha=0.2$)                                             & \textbf{2.90 (-0.95)} \\
                PyramidNet-200 + DCutMix ($K=3, \alpha=\frac{1}{3}$)                                            & \textbf{2.42 (-1.43)} \\
                \bottomrule
                \end{tabular}
    \end{minipage}
    \caption{Comparison of DCutMix and DMixup against other augmentations and regularization methods for PyramidNet-110, 200 models on CIFAR-100 dataset (left) and CIFAR-10 dataset (right). \jhnips{The values in the parentheses of the top-1 error denote the reduced amount of the error compared to the vanilla model, where no augmentation or regularization was applied. }}
    \label{table:comparison_dcutmix_dmixup_cifar}
\end{table}

\begin{table*}[t]
\small
\centering
\tabcolsep=0.1cm 
\begin{tabular}{@{}ccccccccc@{}}
\toprule
Model & Vanilla & Mixup & CutMix & PuzzleMix & Co-Mixup & {StyleCutMix} & DMixup (ours) & DCutMix (ours) \\ \midrule
a & 23.59 (1x) & 22.43 (1.04x) & 21.29 (1.10x) & 20.62 (2.73x) & 19.87 (29.34x) & {20.49 (56.52x)} & 21.68 (1.14x) & 20.50 (1.07x)\\ 
b & 21.70 (1x) & 20.08 (1.02x) & 20.55 (1.04x) & 19.24 (2.25x) & 19.15 (13.46x) & {18.90 (74.98x)} & \textbf{19.13} (1.02x) & \textbf{18.64} (1.01x)  \\
c & 21.79 (1x) & 21.70 (1.02x) & 22.28 (1.05x) & 21.12 (2.22x) & 19.78 (6.93x) & {19.94 (43.48x)} & 20.09 (1.02x) & 20.48 (1.02x) \\
\bottomrule
\end{tabular}
\caption{\small Comparison of Top-1 error rate and training time for \jhnips{the state-of-the-art image mixing augmentations} on CIFAR-100 with various backbone models. (a): PreActResNet18~\cite{he2016identity}, (b): WRN16-8~\cite{zagoruyko2016wide}, (c): ResNeXt29-4-24~\cite{xie2017aggregated}. Values in the parentheses denote the ratio of average training time relative to the baseline (vanilla) case.}
\label{table:compare_sota_mixing_augmentations}
\end{table*}

\section{Experiments and Discussion}
\label{sec:experiment}
\chaaaaai{Here, we experimentally verify the effect of the K-expanded image mixing augmentation. First, we show the improved classification performance after applying our method to CIFAR-10/100  and analyze its advantages in terms of the shape of the loss landscape. Second, on ImageNet, we propose an elaborately designed K-image mixing augmentation that considers the saliency map to overcome label noise. Moreover, we present the experimental result on classification and adversarial robustness for further discussion. Finally, we demonstrate the effectiveness of the proposed data subsampling method and its practical application in NAS.}
\subsection{\chaaaaai{Experimental Result on Image Classification}}

\subsubsection{\chaaaaai{CIFAR-10/100}} We present classification test results on CIFAR-10 and 100~\cite{krizhevsky2009learning} datasets in Table~\ref{table:comparison_dcutmix_dmixup_cifar}.
The results were obtained from the equivalent training and augmentation-specific hyper-parameter setup used in \cite{yun2019cutmix}. 
Firstly, Table \ref{table:comparison_dcutmix_dmixup_cifar} (left) presents the superiority of DCutMix and DMixup over the other augmentation and regularization methods on the CIFAR-100 dataset.
First, for light-weight backbone PyramidNet-110 \cite{han2017deep}, DCutMix and DMixup improve the performance compared to \jhnips{the baselines (i.e., CutMix and Mixup)} by approximately 1\% and 0.32\%, respectively.
For the deeper neural network PyramidNet-200, DCutMix and DMixup achieved the enhancement compared to \jhnips{the baselines}, and DCutMix shows the lowest top-1 error compared to other baselines.
Secondly, we evaluate our proposed methods on the CIFAR-10 dataset as shown in Table \ref{table:comparison_dcutmix_dmixup_cifar} (right). \chaaaaai{We again observe that DCutMix and DMixup both achieved performance enhancement.}
Specifically, DCutMix achieved the best performance among the baseline augmentation methods we tested. 

\paragraph{Comparsion to Recent Augmentation Methods}
\yjr{Also, we further compare our DCutMix and DMixup with state-of-the-art image-mixing augmentation methods, including PuzzleMix~\cite{(PuzzleMix)kim2020puzzle}, Co-Mixup~\cite{(CoMixup)kim2021comixup}, and StyleMix~\cite{hong2021stylemix} , in Table \ref{table:compare_sota_mixing_augmentations}. As seen in the results, DCutMix achieved better accuracy than PuzzleMix and a competitive accuracy compared to the Co-Mixup. \yjr{We note that the proposed augmentation provides comparable classification performance with achieving superior calculation time compared with recently published augmentation methods such as StyleMix and Co-Mixup~\cite{kim2021comixup}.} 
Co-Mixup and StyleMix each require more training time overhead (over 20 times and 50 times) than ours due to the high optimization cost. These overall results demonstrate the effectiveness of the proposed K-image generalization for augmentation methods.}

\begin{figure*}[t]
    \centering
    \begin{subfigure}[t]{0.48\linewidth}
        \includegraphics[width=0.93\linewidth]{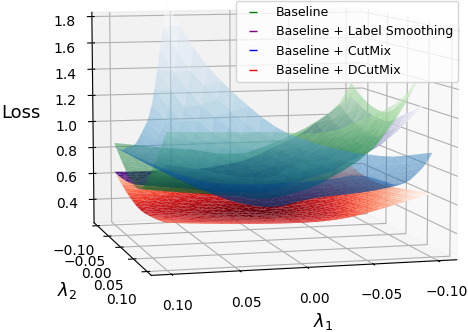}
        \centering
        \caption{\jh{3-d loss surface plot by perturbing the model across the two directions using PyHessian.}}
        \label{fig:wide_local_minima_pyheyssian}
    \end{subfigure}
    \hspace{1mm}
    \begin{subfigure}[t]{0.48\linewidth}
        \includegraphics[width=0.9\linewidth]{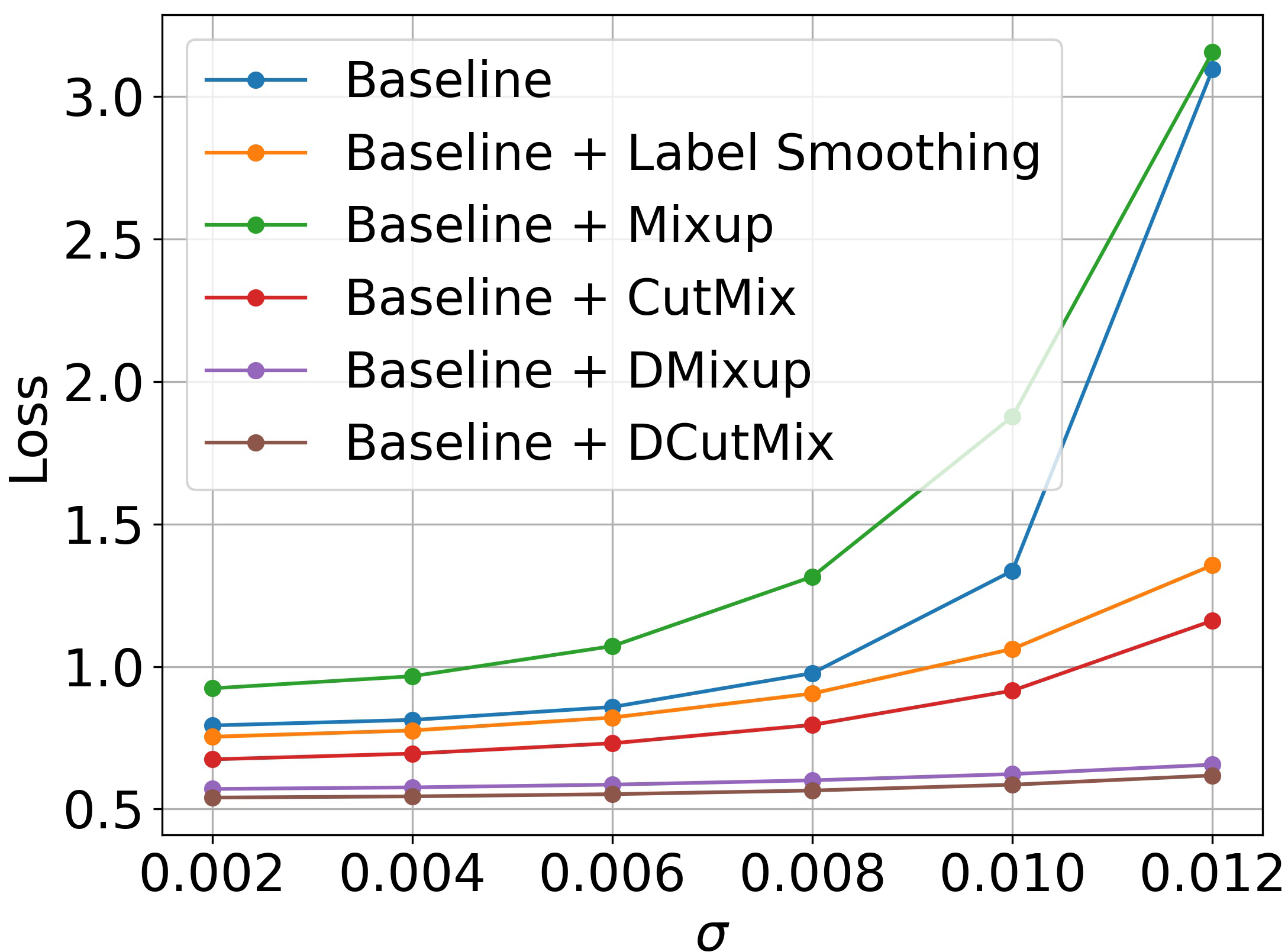}
        \centering
        \caption{\jh{2-d loss surface plot by perturbing the model across one random direction.}}
        \label{fig:wide_local_minima_2d}
    \end{subfigure}
    \caption{Comparison of image mixing augmentation and regularization methods in perspective of the loss-surface near local minima. We measured the loss surfaces on the test set of CIFAR-100 with PyramidNet~\cite{han2017deep}.}
    \label{fig:loss_surface_analysis}
\end{figure*}

\paragraph{\chaaaaai{Analysis on the shape of the loss landscape}} For more explicit investigation, we analyze \jh{DCutMix with} \yjr{regard to} its loss landscape.
\chaaaaai{Flatness of the loss landscape near local minima has been considered as a key indicator of improved model generalization in various situations in numerous previous studies~\cite{keskar2016large, pereyra2017regularizing, zhang2018deep, chaudhari2019entropy, cha2020cpr}.
Regarding the shape of the loss landscape, convergence to a wide (flat) local minima is generally considered to represent a model with better generalization performance on an unseen test dataset.}

\chaaaaai{Accordingly, we use the PyHessian \cite{yao2019pyhessian} \yjr{framework to obtain the loss landscape patterns of each model}, as illustrated in Figure~\ref{fig:wide_local_minima_pyheyssian}.
}
\yjr{The plotted result shows that DCutMix has the widest loss landscape near local minima among the compared models. Moreover, DCutMix exhibits lower losses overall, denoting good generalization to the unseen test data as well. We further plotted the patterns of loss landscape for each model in Figure }\ref{fig:wide_local_minima_2d} \yjr{by perturbing the model parameters with random Gaussian noise through increasing the degree of variance}  $\sigma$~\cite{zhang2018deep}.
DCutMix and DMixup clearly exhibited the widest and lowest loss landscape compared to the other methods, including CutMix and Mixup, which are baseline two-image mixing augmentation methods. For DMixup particularly, we observed that the convergence stability was better than that of Mixup. We believe that this reveals the superiority of the proposed K-image mixing augmentation.

From a more analytical point of view, we can hypothesize that the K-image generalization of DCutMix and DMixup, regarding the wide flat local minima, is attributable to their labels being softer than that of CutMix and Mixup. Several researchers have reported that a model trained with an artificially smoothed label can result in the model converging to wide local minima, thus achieving better generalization~\cite{pereyra2017regularizing, zhang2018deep, cha2020cpr}, \yjr{as in the case of the superior results} of Label Smoothing~\cite{pereyra2017regularizing} compared to \jhnips{the baseline} in Figure~\ref{fig:wide_local_minima_pyheyssian} and Figure \ref{fig:wide_local_minima_2d}.
However, \chaaaaai{as opposed to the previous regularization methods using \textit{artificially} smoothed label}, note that our softened label \textcolor{black}{\textit{directly}} reveals the augmented ratio of several images, and hence, we conjecture the tendency can be a key factor why the model trained by our approach converges to lower and \jhnips{wider} local minima.

\subsubsection{\jhnips{ImageNet}}
\begin{figure*}[t]
    \centering
    \begin{subfigure}[t]{0.495\linewidth}
        \includegraphics[width=0.99\linewidth]{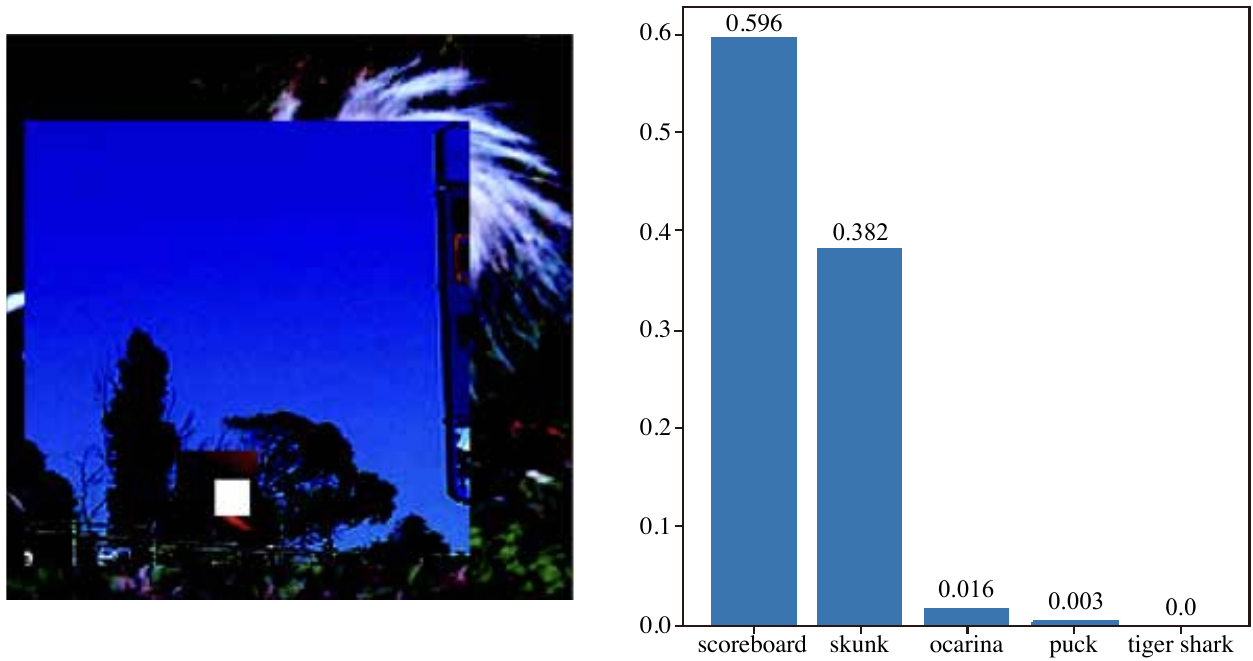}
        \centering
    \end{subfigure}
    \begin{subfigure}[t]{0.495\linewidth}
        \includegraphics[width=0.99\linewidth]{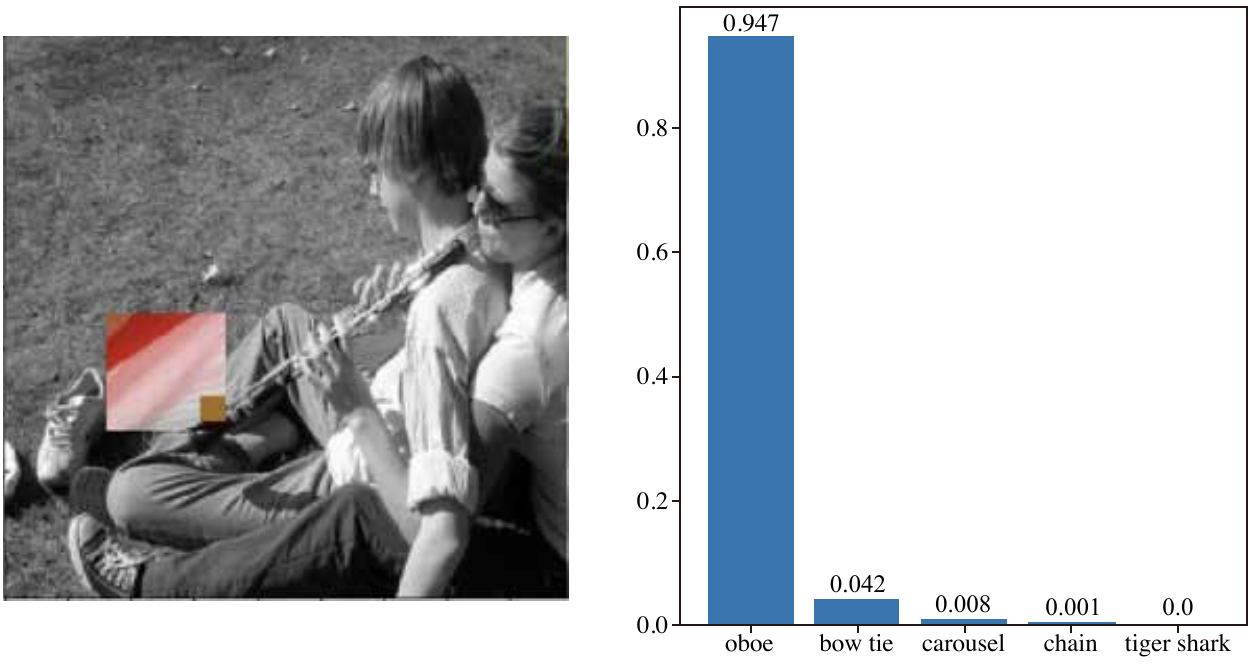}
        \centering
    \end{subfigure}
    \begin{subfigure}[t]{0.495\linewidth}
        \includegraphics[width=0.99\linewidth]{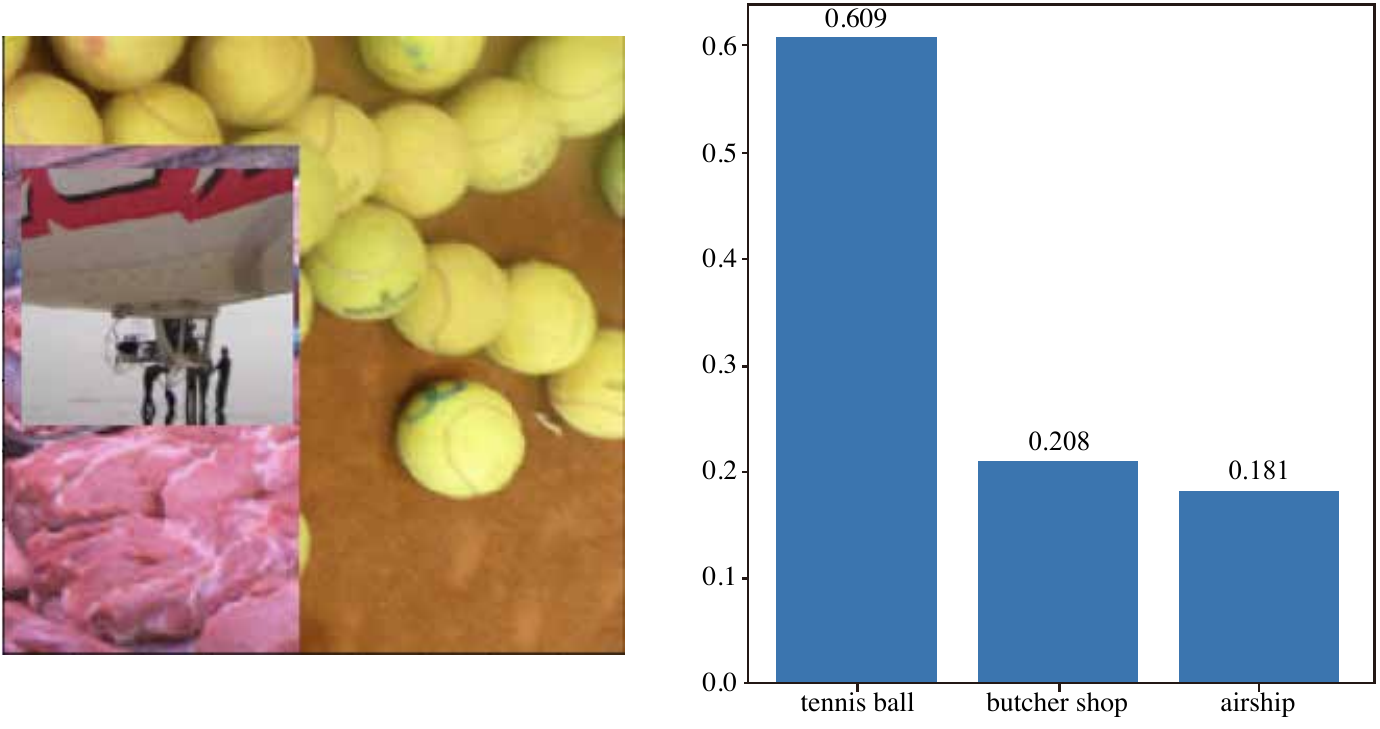}
        \centering
    \end{subfigure}
    \begin{subfigure}[t]{0.495\linewidth}
        \includegraphics[width=0.99\linewidth]{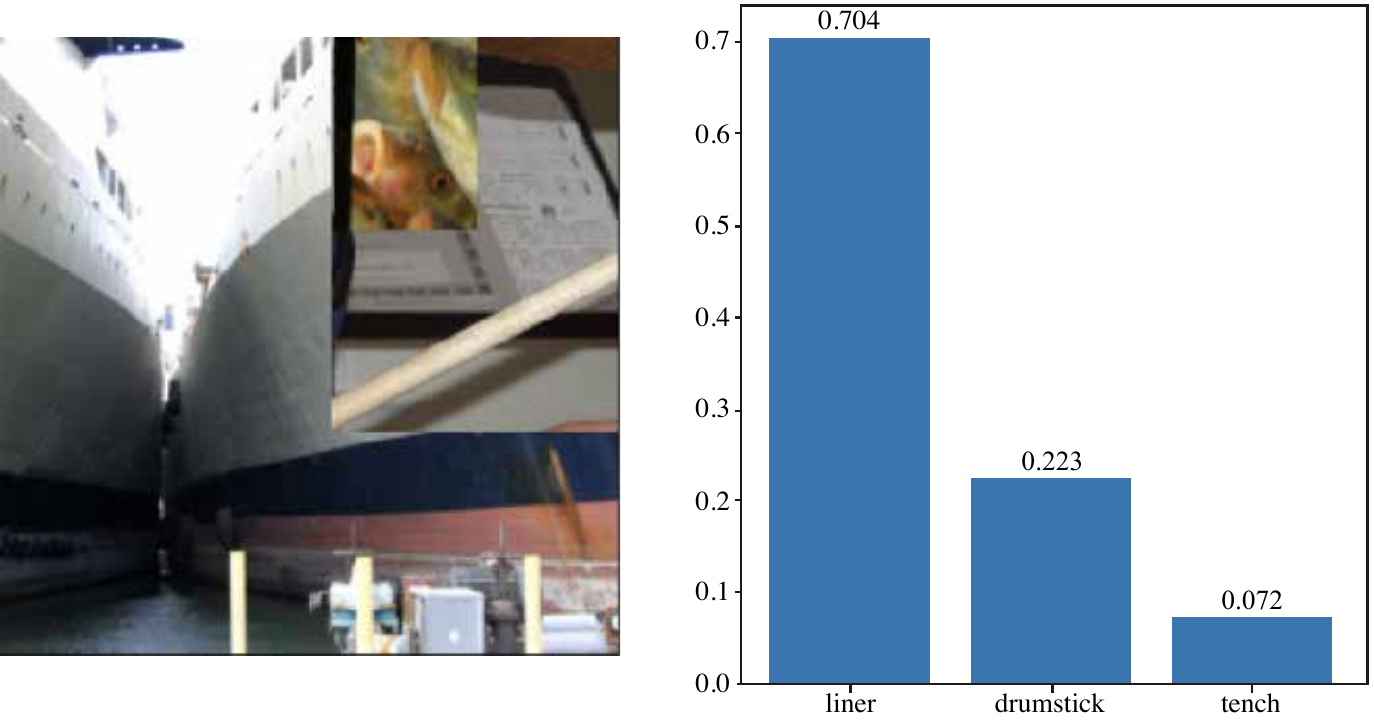}
        \centering
    \end{subfigure}
    \caption{\jhnips{Qualitative example of an augmented image and its soft label distribution for top K classes (top: DCutMix, bottom: Saliency-DCutMix). Note that for the DCutMix case, visually salient information of top K classes is not fully contained (e.g., "Scoreboard" or "bow-tie") in the augmented image. Meanwhile, visually salient information of all the top K classes is fully contained for the Saliency-DCutMix case.}}
    \label{fig:qualitative_examples_dcutmix_saliencydcutmix}
\end{figure*}

\jhnips{
We present ImageNet classification results of DCutMix and DMixup compared to the two-image mixing baselines: CutMix and Mixup. The results are obtained under the equivalent training and augmentation-specific hyper-parameter setup used in~\cite{yun2019cutmix}. 
As presented in Table \ref{table:comparison_dcutmix_dmixup_imagenet} (left), DMixup considerably improved the performance of Mixup and Manifold Mixup by 0.7\% and 0.62\% respectively, while reducing the top-1 error. However, DCutMix exhibited a higher top-1 error rate compared to CutMix. This result was attributable to DCutMix suffering from the label noise problem,  where a background object other than the ground truth class object is contained in the randomly cropped image~\cite{yun2021re}, as shown in Figure \ref{fig:qualitative_examples_dcutmix_saliencydcutmix}.
\yjr{Moreover, Table \ref{table:comparison_dcutmix_dmixup_imagenet} (right) reveals that as the number of mixing images $K$ increased, the performance of DCutMix deteriorated due to the higher probability of background objects being accumulated.}}

\begin{table}[t]
\begin{minipage}{.5\textwidth}
        \small
        \centering
        \tabcolsep=0.05cm
            \resizebox{1\linewidth}{!}{%
                \begin{tabular}{@{}lcccc@{}}
                \toprule
                \multirow{2}{*}{Model}            & \multirow{2}{*}{\begin{tabular}[c]{@{}c@{}}Top-1 \\ Err (\%)\end{tabular}} & \multirow{2}{*}{\begin{tabular}[c]{@{}c@{}}Top-5 \\ Err (\%)\end{tabular}} & \\
                                                  &    \\ \midrule
                ResNet-50 (Baseline)                               & 23.68 & 7.05 \\
                ResNet-50 + Mixup                    & 22.58 & 6.40 \\
                ResNet-50 + Manifold Mixup                    & 22.50 & 6.21 \\
                ResNet-50 + CutMix                    & 21.40 & 5.92 \\
                ResNet-50 + DMixup ($K=3, \alpha=\frac{1}{3}$)                   & 21.88 & 6.15 \\
                ResNet-50 + DCutMix ($K=3, \alpha=\frac{1}{3}$)                   & 21.76 & 5.91 \\
                ResNet-50 + Saliency-DCutMix ($K=3, \alpha=\frac{1}{3}$)                   & 21.38 & 5.87 \\
                \bottomrule
                \end{tabular}
            }
    \end{minipage}
    \begin{minipage}{.5\textwidth}
        \small
        \centering
        \tabcolsep=0.05cm
            \resizebox{1\linewidth}{!}{%
                \begin{tabular}{@{}lcccccc@{}}
                \toprule
                \multirow{2}{*}{Method}            & \multirow{2}{*}{$\alpha$}  &
                \multirow{2}{*}{K}            &\multirow{2}{*}{\begin{tabular}[c]{@{}c@{}}Top-1 \\ Err (\%)\end{tabular}} & \multirow{2}{*}{\begin{tabular}[c]{@{}c@{}}Top-5 \\ Err (\%)\end{tabular}} & \\
                                                  &    \\ \midrule

                ResNet-50 + DCutMix & 1 & 3 & 21.62 & 5.92 \\
                ResNet-50 + DCutMix & 1 & 4 & 22.17 & 6.23 \\
                ResNet-50 + DCutMix & 1 & 5 & 22.02 & 6.05 \\
                ResNet-50 + DCutMix & 1 & 7 & 22.39 & 6.17 \\
                ResNet-50 + DCutMix & 1 & 9 & 22.58 & 6.34 \\
                ResNet-50 + Saliency-DCutMix & 1 & 3 & 21.40 & 5.87 \\
                ResNet-50 + Saliency-DCutMix & 1 & 4 & 21.55 & 6.00 \\
                ResNet-50 + Saliency-DCutMix & 1 & 5 & 21.70 & 5.93 \\
                ResNet-50 + Saliency-DCutMix & 1 & 7 & 21.44 & 5.93 \\
                \bottomrule
                \end{tabular}
            }
    \end{minipage}

\caption{\jhnips{Performance of DMixup and DCutMix, and Saliency-DCutMix on ImageNet (left). Impact of $K$ on DCutMix and Saliency-DCutMix (right).}}

\label{table:comparison_dcutmix_dmixup_imagenet}
\end{table}

\jhnips{
\yjr{To address this label noise problem}, we devised a more sophisticated mixing method named Saliency-DCutMix, which employs saliency-map information for integration with our DCutMix. First, we obtain a salient image patch by selecting the most salient pixel point of the saliency map as the center point, as suggested in~\cite{(SaliencyMix)uddin2021saliencymix}. Here, the width and height of each patch from (\ref{eq:stick_breaking}) to ensure the Dirichlet distribution is followed. Consequently, we mixed these salient image patches with SBP, similar to in DCutMix, as given in (\ref{eq:cut_mix_sbp}). Figure \ref{fig:qualitative_examples_dcutmix_saliencydcutmix} shows the qualitative examples of DCutMix and Saliency-DCutMix. Samples augmented with DCutMix contain background class objects other than the ground truth object, which could lead to label noise during training. Meanwhile, samples augmented with Saliency-DCutMix reveal that the foreground class objects are mixed without background class objects being included. In Table  \ref{table:comparison_dcutmix_dmixup_imagenet} (right), Saliency-DCutMix indeed exhibits relatively stable and significantly improved performance regardless of K compared to DCutMix. Furthermore, Saliency-DCutMix achieved higher performance than its baseline two-image mixing augmentation method, CutMix, as demonstrated in Table~ \ref{table:comparison_dcutmix_dmixup_imagenet} (left).
}

\paragraph{ImageNet-O}

To evaluate the robustness of our proposed model to the out-of-distribution (OOD) data samples, we performed tests on the ImageNet-O dataset \cite{hendrycks2021natural}. 
The dataset contains OOD images whose class labels do not belong to 1000 classes of the ImageNet-1K dataset. The most ideal output of a classification model against the OOD data sample is uniformly predicting all classes with low confidence because a class in the OOD dataset was not considered when training the classification model. 
These OOD images reliably cause various models to be misclassified with high confidence. To evaluate the robustness of each model against the OOD dataset samples, we measured the area under the precision-recall curve (AUPR) on the ImageNet-O dataset, where a higher AUPR denotes that the model robustly predicted OOD samples with lower confidence. Notably, In Table \ref{table:comparison_imagenet_a_o}, the model trained without augmentation (Vanilla) exhibited the best AUPR. All the augmentation methods are highly over-confident for the OOD samples, and the results demonstrate the fragility of the augmentation methods when a label distribution shift is present.

\begin{table}[t]
\small
\centering
\tabcolsep=0.2cm
\begin{tabular}{@{}lccc@{}}
\toprule
\multirow{2}{*}{Model}            & \multirow{2}{*}{\begin{tabular}[c]{@{}c@{}}ImageNet-O \\ AUPR (\%)\end{tabular}} & \\
                                  &    \\ \midrule
                                  
Vanilla                               & 16.96 \\
Mixup                    & 16.30 \\
DMixup     & 16.87 \\
CutMix                    & 15.85 \\
DCutMix & 16.73 \\
Saliency-DCutMix                   & 16.00 \\

\bottomrule
\end{tabular}

\caption{\jhnips{Performance of various augmentation methods on ImageNet-O. We used ResNet-50 for all the methods. Vanilla denotes ResNet-50 trained without augmentation.}}
\label{table:comparison_imagenet_a_o}
\end{table}

\paragraph{\cha{Adversarial robustness}} 
\chaaaaai{After the vulnerability of deep neural networks was elucidated by \cite{(vurlnerability)szegedy2013intriguing}, achieving superior classification performance for both non-attacked examples (standard accuracy) and adversarial robustness (robust accuracy) has been considered as the key factor for making the deep neural network truly robust and reliable \cite{(FGSM)goodfellow2014explaining, (PGD)madry2017towards, (input_transform)guo2017countering, (benchmarking)dong2020benchmarking}.
To achieve higher adversarial robustness, several competing attack and defense methods have been alternately proposed  \cite{(FGSM)goodfellow2014explaining, (PGD)madry2017towards, (input_transform)guo2017countering, (benchmarking)dong2020benchmarking}.
\jhnips{In contrast to the above-mentioned studies, \cite{(PGD)madry2017towards, yun2019cutmix} have reported that training a model with an input transformation or augmentation enhances the robustness of the model against adversarial examples without adversarial training \cite{(PGD)madry2017towards}, which suffers from high training cost and a severe trade-off between standard and robust accuracy.}}

\begin{table*}[ht]
\centering
\small
\noindent
\begin{tabular}{c||c|c|c|c}
\hline
\multirow{2}{*}{\begin{tabular}[c]{@{}c@{}}\\ Model\end{tabular}} & White-box                                                               & Gray-box                                                               & Black-box                      & \multirow{2}{*}{\begin{tabular}[c]{@{}c@{}}\\ ImageNet-A\end{tabular}}                                        \\ \cline{2-4} 
                                                                    & \begin{tabular}[c]{@{}c@{}}FGSM\\ ($L_\infty, \epsilon=8$)\end{tabular} & \begin{tabular}[c]{@{}c@{}}PGD\\ ($L_\infty, \epsilon=8$)\end{tabular} & \begin{tabular}[c]{@{}c@{}}PGD\\ ($L_\infty, \epsilon=8$)\end{tabular} \\ \hline \hline
\begin{tabular}[c]{@{}c@{}}ResNet-50 (Baseline)\end{tabular}      & 14.34                                                                   & 40.74                                                                  & 46.01  & 3.38                                                                \\ \hline 
\begin{tabular}[c]{@{}c@{}}+ CutMix \\ \end{tabular}                                                            & 40.89 (+26.55)                                                          & 43.38 (+2.64)                                                          & 51.06 (+5.05)   &  7.46 (+4.08)                                                     \\ \hline
\begin{tabular}[c]{@{}c@{}}+ DCutMix\\ ($K=3, \alpha=\frac{1}{3}$)\end{tabular}                                                           & 32.90 (+18.56)                                                          & 43.63 (+2.89)                                                          & 51.73 (+5.72)     & 5.84 (+2.46)                                                    \\ \hline
\begin{tabular}[c]{@{}c@{}}+ Saliency-DCutMix\\($K=3, \alpha=\frac{1}{3}$)\end{tabular}       & \textbf{41.37 (+27.03)}                                                          & \textbf{44.27 (+3.53)}                                                          & \textbf{51.91 (+5.90)}    & \textbf{7.69 (+4.31)}                                                      \\ \hline
\end{tabular}
\caption{\jhnips{Top-1 robust accuracy on ImageNet against various adversarial attacks (2nd, 3rd, 4th column) and accuracy on ImageNet-A (last column).} The values in parentheses denote accuracy increment compared to the Baseline case.}
\label{table:adversarial}
\end{table*}

\begin{figure*}[t]
    \centering
    \begin{subfigure}[t]{0.485\linewidth}
        \includegraphics[width=0.75\linewidth]{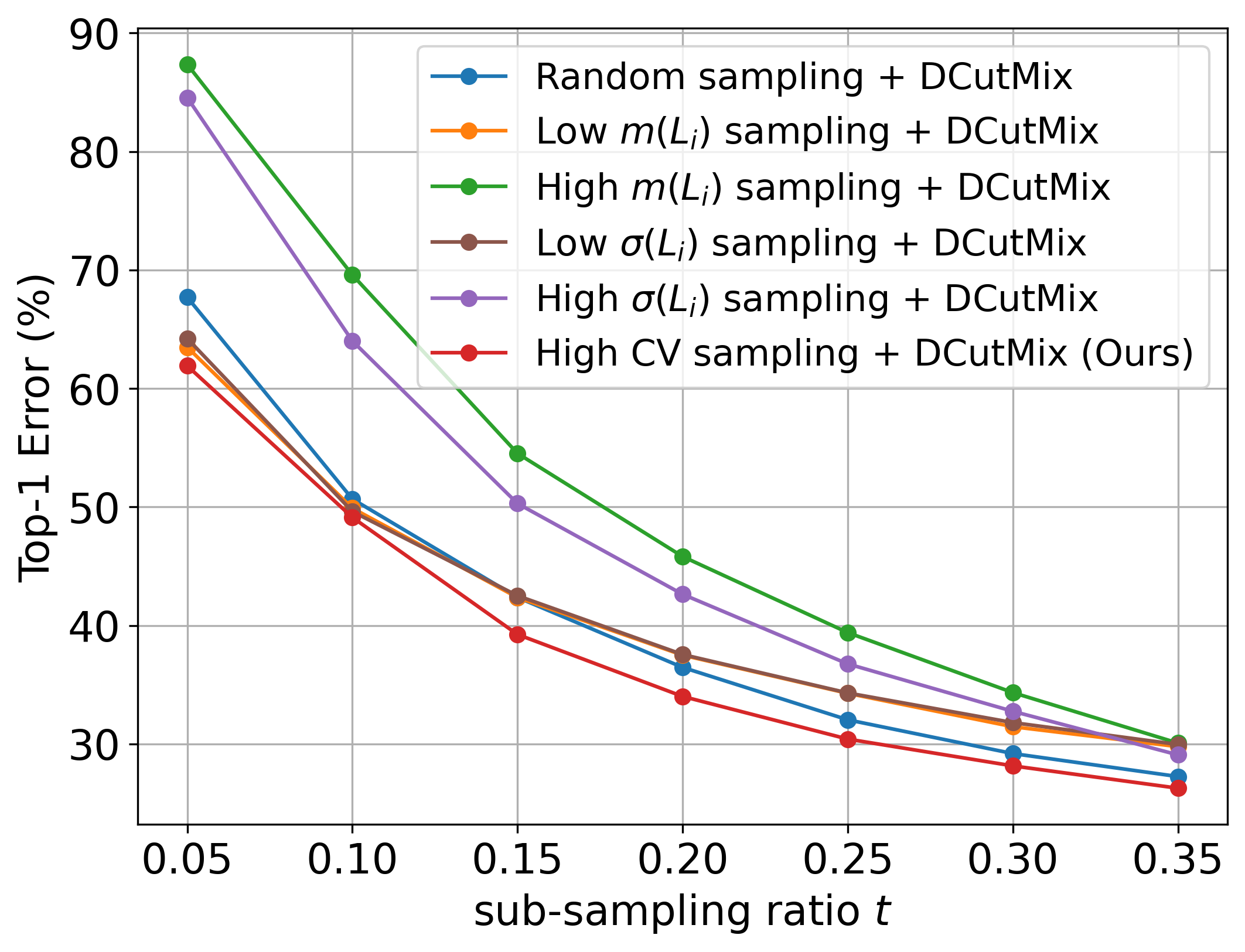}
        \centering
        \caption{Top-1 error plotting for data subsampling methods. $t$ denotes the ratio of images in the subsampled set to those in the whole training set.}
        \label{fig:various_sampling_methods}
    \end{subfigure}
    \hspace{1mm}
    \begin{subfigure}[t]{0.485\linewidth}
        \includegraphics[width=1\linewidth]{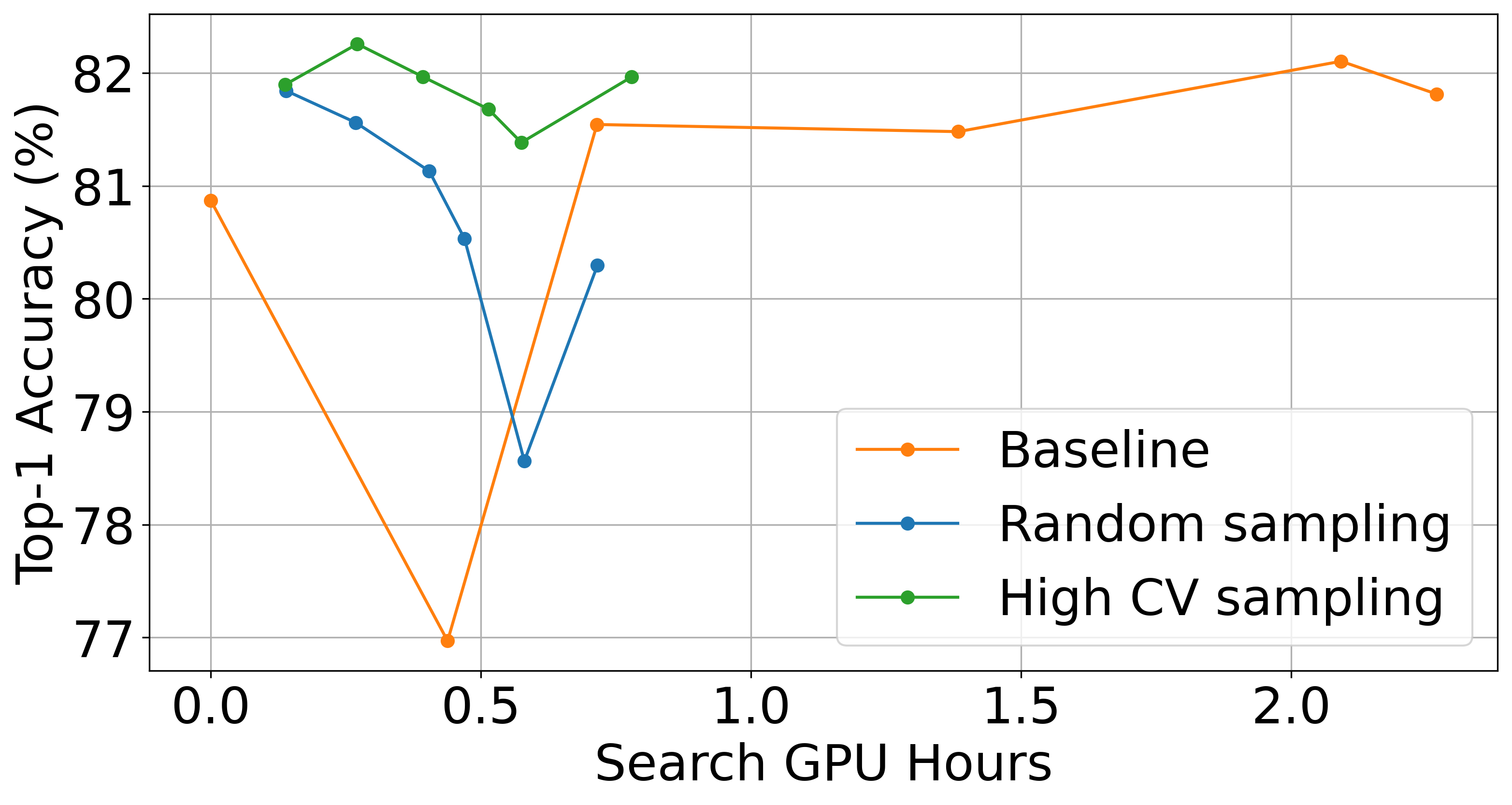}
        \centering
        \caption{Top-1 accuracy plotting of NAS given GPU searching cost on CIFAR-100.}
        \label{fig:cifar100_nas}
    \end{subfigure}
    \caption{Experiments on data subsampling: performance of various data subsampling methods (a). Graph (b) shows the speedup of NAS from the proposed subsampling method: searching on the entire CIFAR-100 dataset (Baseline), on randomly subsampled CIFAR-100, and on subsampled CIFAR-100 by high CV measure.}
    \label{fig:ablation_study_on_selection_function}
\end{figure*} 

In this section, we demonstrate the additional advantage of K-image mixing augmentation in terms of adversarial robustness.
. We selected each classification model (ResNet-50) trained by the baseline (w/o augmentation), CutMix, DCutMix, and Saliency-DCutMix using the ImageNet training dataset. To evaluate adversarial robustness against more diverse types of attack, we considered not only white-box attacks as in  \cite{yun2019cutmix}, but also gray- and black-box attacks \cite{(input_transform)guo2017countering}.

Regarding white-box attack where the attacker can freely access to the model's parameter, we use FGSM $\ell_{\infty}$ attack~\cite{(FGSM)goodfellow2014explaining} with $\epsilon=8$, as in \cite{yun2019cutmix}.
The black-box attack is more challenging for an attacker because they do not have any information about the target model to be attacked. For this case, we set a substitute model (\textit{i.e.} ResNet152) and made adversarial examples by attacking it.
Gray-box attack is a compromise between white- and black-box attacks: the attacker knows the architecture of the model (\textit{i.e.} ResNet50)) without having access to the weight parameters.
Therefore, we generate adversarial examples using a substitute ResNet50 model trained with the ImageNet dataset using a different random seed.
For gray- and black-box attacks, we generated adversarial examples of ImageNet validation dataset using a more strong attack method called as PGD $\ell_{\infty}$ attack~\cite{(PGD)madry2017towards} with epsilon $\epsilon=8$.

Table \ref{table:adversarial} shows top-1 accuracy on given adversarial examples generated by each attack.
As proposed in \cite{yun2019cutmix}, a model trained with CutMix exhibits better adversarial robustness than the baseline case against all types of attacks. DCutMix achieves more improved adversarial robustness against gray- and black-box attacks compared to CutMix. However, this was not the case against the white-box attack. Notably, the white-box attack is the most powerful attack among the attacks. We hypothesized that the label noise problem associated with DCutMix degrades adversarial robustness against a strong attack (white-box), and this hypothesis was indirectly confirmed through experiments on Saliency-DCutMix. We observed that saliency-map-guided DCutmix (Saliency-DCutmix) has stronger adversarial robustness than CutMix and DCutMix for all types of attacks.

For a more sophisticated investigation of robustness against shifts in input data distribution, we evaluated the accuracy of augmentation methods on the ImageNet-A dataset \cite{hendrycks2021natural}. this dataset contains natural adversarial examples, which cause the wrong classification in an ImageNet-pre-trained model without any adversarial attacks. In Table \ref{table:adversarial}, we found that the performance tendency on ImageNet-A is similar to that of adversarially attacked ImageNet. CutMix exhibited better accuracy than the baseline case and DCutMix. Meanwhile, Saliency-DCutMix improved the CutMix accuracy, exhibiting better generalization on natural adversarial examples and, hence, better robustness on the input data distribution shift.

\begin{table*}[h]
  \centering
      \begin{tabular}{@{}lcccccc@{}}
        \toprule
        \multirow{2}{*}{\textbf{Architecture}}  & \multirow{2}{*}{\begin{tabular}[c]{@{}c@{}}\textbf{Top-1} \\ \textbf{Err (\%)}\end{tabular}} & \multirow{2}{*}{\begin{tabular}[c]{@{}c@{}}\textbf{Top-5} \\ \textbf{Err (\%)}\end{tabular}} & \multirow{2}{*}{\begin{tabular}[c]{@{}c@{}}\textbf{\# Params} \\ \textbf{(M)}\end{tabular}} & \multirow{2}{*}{\begin{tabular}[c]{@{}c@{}}\textbf{\# FLOPs} \\ \textbf{(M)}\end{tabular}} & \multirow{2}{*}{\begin{tabular}[c]{@{}c@{}}\textbf{Search Cost} \\ \textbf{(GPU days)} \end{tabular}} &  \multirow{2}{*}{\begin{tabular}[c]{@{}c@{}}\textbf{Search} \\ \textbf{method} \end{tabular}}\\
                        &               &               &       &         &           \\
    \midrule
    AmoebaNet-C~\cite{real2018regularized}        & 24.3 & 7.6  & 6.4 & 570  & 3150 & evolution \\
    MnasNet-92~\cite{tan2018mnasnet}              & 25.2 & 8.0  & 4.4 & 388  & -    & RL \\
    ProxylessNAS ~\cite{cai2018proxylessnas}       & 24.9 & 7.5  & 7.1 & 465  & 8.3  & gradient-based \\
    SNAS~\cite{xie2018snas}     & 27.3 & 9.2  & 4.3 & 522  & 1.5  & gradient-based \\
    BayesNAS~\cite{zhou2019bayesnas}                         & 26.5 & 8.9  & 3.9 & -  & 0.2  & gradient-based \\
    \midrule
    PC-DARTS (CIFAR10) \cite{xu2019pc}               & 25.1 & 7.8  & 5.3 & 586  & 0.1  & gradient-based \\
    PC-DARTS (ImageNet) \cite{xu2019pc}       & 24.2 & 7.3  & 5.3 & 597  & 3.8 & gradient-based \\
    PC-DARTS (CIFAR100) \cite{xu2019pc}       & 23.8 & 7.09  & 6.3 & 730  & 0.1 & gradient-based \\
    PC-DARTS (High CV Sub-sampled CIFAR100)        & 24.3 & 7.2  & 5.8 & 671  & \textbf{0.01} & gradient-based \\
    \bottomrule
      \end{tabular}
  
  \caption{Comparison of the state-of-the-art NAS methods on ImageNet under comparably small resource constraints. ($\cdot$) denotes the proxy dataset where the architecture was searched on.}
  \label{table:imagenet_nas}
\end{table*}

\subsection{Data Subsampling and Application}
\label{sec:experiment_data_subsampling}
\subsubsection{Data Subsampling} We investigate the effect of the proposed subsampling method when trained with DCutMix as an augmentation in Figure \ref{fig:various_sampling_methods}. In the figure, we compare our data subsampling method with others using different subsampling measures. For all subsequent experiments involving data subsampling, a full 10K CIFAR-100 validation set was used for evaluation, and we reported the averaged results for three independent random seeds using PyramidNet  \cite{han2017deep}.

As shown in Figure \ref{fig:various_sampling_methods}, sampling the easy-only or hard-only examples based on $m(L_i)$ shows deteriorated performance compared to the random subsampling. 
The hard-only subsampling severely suffered from poor generalization. \jhnips{This result indicates that subsampling only hard samples where the salient regions were occluded by image-mixing augmentation being applied (see Figure \ref{fig:sampling_toy_examples}) is not desirable under the constraint of a small number of training samples.
In a similar manner, subsampling only easy samples extract the biased data samples that cannot be helpful for better generalization.}
\yjr{Moreover}, simply employing standard deviation $\sigma(L_i)$ as subsampling measure \yjr{induced} a similar test error plot as that from the above mean-based sampling methods. On the other hand, our high-CV-based subsampling significantly outperformed the random sampling. Specifically, the test error was 5.79\% lower when the number of subsampled training samples was extremely small (i.e., $t=0.05$). High CV subsampling enables us to acquire various levels of data samples, from easy to hard. High CV subsampling basically selects the easy data samples, which can frequently become hard samples depending on the image-mixing augmentation. Therefore, High-CV subsampling leads to better performance when training with image-mixing augmentation. We demonstrated the superiority of our subsampling method over other subsampling methods employing uncertainty derived by weight dropout \cite{gal2017deep} and \textit{K-Center Coreset} sampling \cite{sener2017active}.

\subsubsection{Application on NAS} 
We further demonstrate the practicality and effectiveness of our proposed data subsampling method on another domain, namely, NAS. Our goal is to reduce the time spent searching the architectures by searching on the subsampled dataset drawn from our framework rather than on the full training dataset.  
\yjr{We demonstrated that the architecture search time is greatly reduced without accuracy degeneration. Notably, the data subset subsampled by our algorithm can be applied to any neural architecture search framework, including gradient-based and non-gradient-based search methods. We adopt one of the most computationally efficient and stabilized NAS methods, PC-DARTS~\cite{xu2019pc}, as our baseline.}

For the searching process, we divided the subsampled (or entire) training dataset into two equal parts, with one for optimizing the network parameters and the other one for optimizing the architecture hyperparameters (i.e., $\alpha, \beta$ in \cite{xu2019pc}). Additionally, we adopted the warm-up strategy during the search process, where only network parameters are optimized. We freeze the hyperparameters  $\alpha, \beta$ for the first 15 epochs as in \cite{xu2019pc}. We applied the warm-up strategy for the first five epochs for the baseline method (i.e., searching on the entire dataset) where the number of total searching epochs was 10 (i.e., the left-most point for the Baseline in Figure \ref{fig:cifar100_nas} of the manuscript). We used Tesla V100 GPU to perform the search.

For the evaluation involving training the searched network from scratch, we used the equivalent training hyperparameters as in \cite{xu2019pc}.
The performance of the neural networks searched on the entire CIFAR-100 dataset (baseline) is plotted in Figure ~\ref{fig:cifar100_nas}. We plotted the performance of neural networks searched on the entire CIFAR-100 dataset (baseline) by adjusting the searching epochs while adjusting the subsampling ratio $t$ for searching on the randomly subsampled dataset and our subsampled dataset drawn by high CV measure. The searching epochs were adjusted while adjusting the subsampling ratio $t$ for searching on the randomly subsampled dataset and our subsampled dataset drawn using a high CV measure. The results demonstrate the outstanding efficiency of our subsampling framework in terms of search time and accuracy. Specifically, it (searching on the high-CV subsampled dataset) achieved comparable accuracy with a 7.7-fold reduction in search time compared to other baselines. Furthermore, it consistently outperformed random subsampling given an equivalent number of data samples for searching.

As listed in Table~\ref{table:imagenet_nas}, we observed that our framework serves as an effective proxy dataset, and the neural network searched using it is well-generalized on ImageNet. Notably, it reduced the GPU search time to as much as 0.01 d (i.e., 16 min) while achieving comparable or even higher accuracy than that of the models searched with PC-DARTS on the entire CIFAR-10, CIFAR-100, and randomly subsampled ImageNet datasets. Moreover, compared to the other NAS methods, ours achieved the best accuracy and significantly lower search computational cost.

\section{Concluding Remarks and Limitation}
\label{sec:conclusion}
In this study, we present the advantages of expanding the number of images for image-mixing augmentation based on various experimental results and analyses. First, we propose the generalized method for K-image mixing augmentation motivated by SBP. Second, we demonstrate that the proposed K-image mixing augmentation improves classification performance. Moreover, from a novel perspective, we demonstrated that the key factor behind this improvement is the convergence to wide local minima. Moreover, we empirically found that increasing the number of images for the image mixing augmentation enhances the adversarial robustness of a classification model against various types of adversarial examples. Additionally, we derived a new subsampling method that utilizes the proposed K-image mixing augmentation in a novel way. We experimentally demonstrate that the proposed subsampling method can effectively reduce search time without performance degradation. We believe our observations can inspire new research directions for image mixing augmentation and data subsampling.

\yjr{\textbf{Limitation: }Because our method focuses on setting a probabilistic framework explaining the CutMix augmentation and its potential effectiveness, we did not employ other semantic knowledge, such as spatial attention or saliency map. However, if strictly targeting the SOTA classification performance, it would be a promising future direction to employ the additional information in the augmentation process. Furthermore, employing the idea in other computer vision tasks, such as object detection and segmentation, will enhance the applicability of the method.}

\section{Acknowledgement}
This work was partly supported by Clova, NAVER corp and the Institute of Information \& communications Technology Planning \& Evaluation (IITP) grant funded by the Korean government(MSIT) (2021-0-01341, Artificial Intelligence Graduate School Program(Chung-Ang University); 2021-0-02067, Next Generation AI for Multi-purpose Video Search).

{\small
\bibliographystyle{IEEEtran}
\bibliography{egbib}
}

\begin{IEEEbiography}[{\includegraphics[width=1in,height=1.25in,clip,keepaspectratio]{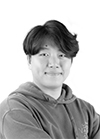}}]
{Joonhyun Jeong} received his Bachelor's degree from the Department of Computer Science and Engineering, Kyung Hee University, South Korea, in 2019. Currently, he is working at NAVER CLOVA, South Korea. His research interests include model compression and data augmentation in computer vision tasks.
\end{IEEEbiography}

\begin{IEEEbiography}[{\includegraphics[width=1in,height=1.25in,clip,keepaspectratio]{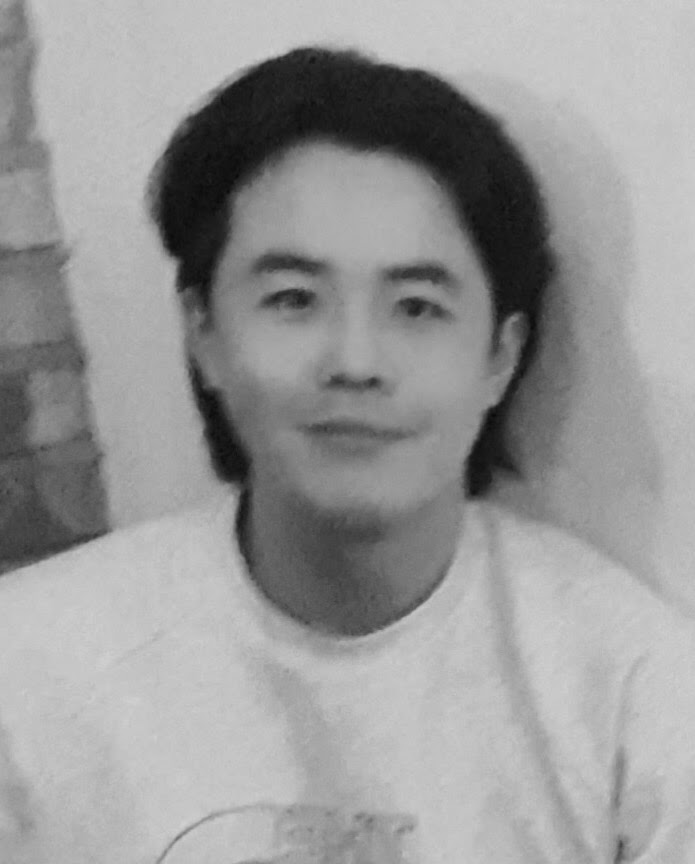}}]
{Sungmin Cha} is currently pursuing a Ph.D. degree in electrical and computer engineering at Seoul National University (SNU) in Seoul, South Korea. He received a B.S degree in computer engineering from Pukyung National University in Busan, South Korea in 2016 and an M.S degree in information and communication engineering from Daegu-Gyeongbuk Institute of Science and Technology (DGIST), in Daegu, South Korea in 2018. His current research interests include deep neural network-based unsupervised image denoising, continual learning, and adversarial robustness for the deep neural networks.
\end{IEEEbiography}

\begin{IEEEbiography}[{\includegraphics[width=1in,height=1.25in,clip,keepaspectratio]{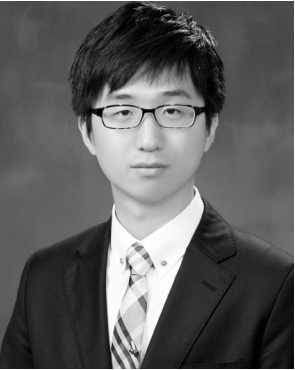}}]{Jongwon Choi} received the B.S. and M.S. degrees in electrical engineering from KAIST, Daejeon, South Korea, in 2012 and 2014, respectively, and the Ph.D. degree in electrical engineering from Seoul National University, Seoul, South Korea, in 2018.
From 2018 to 2020, he was with the Research Intelligence Research Center, Samsung SDS, Seoul, as a Research Engineer. 
In 2020, he joined the Department of Advanced Imaging, Chung-Ang University, Seoul, where he is currently working as an Assistant Professor.
His research interests include the surveillance system with deep learning, the architecture of deep learning, and low-level computer vision algorithms.
\end{IEEEbiography}

\begin{IEEEbiography}[{\includegraphics[width=1in,height=1.25in,clip,keepaspectratio]{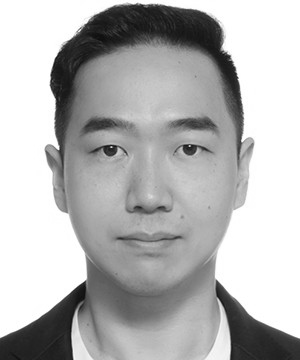}}]
{Sangdoo Yun} received the B.S, M.S, and Ph.D. degrees in electrical engineering and computer science from Seoul National University,
Seoul, South Korea, in 2010, 2013, and 2017, respectively. He is currently a research scientist at NAVER AI LAB. His current research
interests include computer vision, deep learning, and image classification.
\end{IEEEbiography}

\begin{IEEEbiography}[{\includegraphics[width=1in,height=1.25in,clip,keepaspectratio]{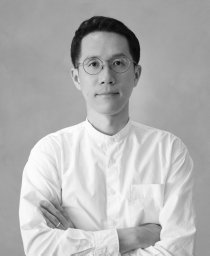}}]{Taesup Moon} received the B.S. degree in electrical engineering from Seoul National University, Seoul, South Korea, in 2002, and the M.S. and Ph.D. degrees in electrical engineering from Stanford University, Stanford, CA, USA, in 2004 and 2008, respectively. From 2008 to 2012, he was a Scientist with Yahoo! Labs, Sunnyvale, CA, USA. He was a Postdoctoral Researcher with the Department of Statistics, UC Berkeley, from 2012 to 2013. From 2013 to 2015, he was a Research Staff Member with the Samsung Advanced Institute of Technology (SAIT), and from 2015 to 2017, he was an Assistant Professor with the Department of Information and Communication Engineering, Daegu Gyeongbuk Institute of Science and Technology (DGIST), and from 2017 to 2021, he was an Associate Professor with the Department of Electrical and Computer Engineering, Sungkyunkwan University (SKKU), Suwon, South Korea. He is currently an Associate Professor at the Department of Electrical and Computer Engineering, Seoul National University (SNU), Seoul, South Korea. His current research interests include machine/deep learning, signal processing, information theory, and various (big) data science applications
\end{IEEEbiography}

\begin{IEEEbiography}[{\includegraphics[width=1in,height=1.25in,clip,keepaspectratio]{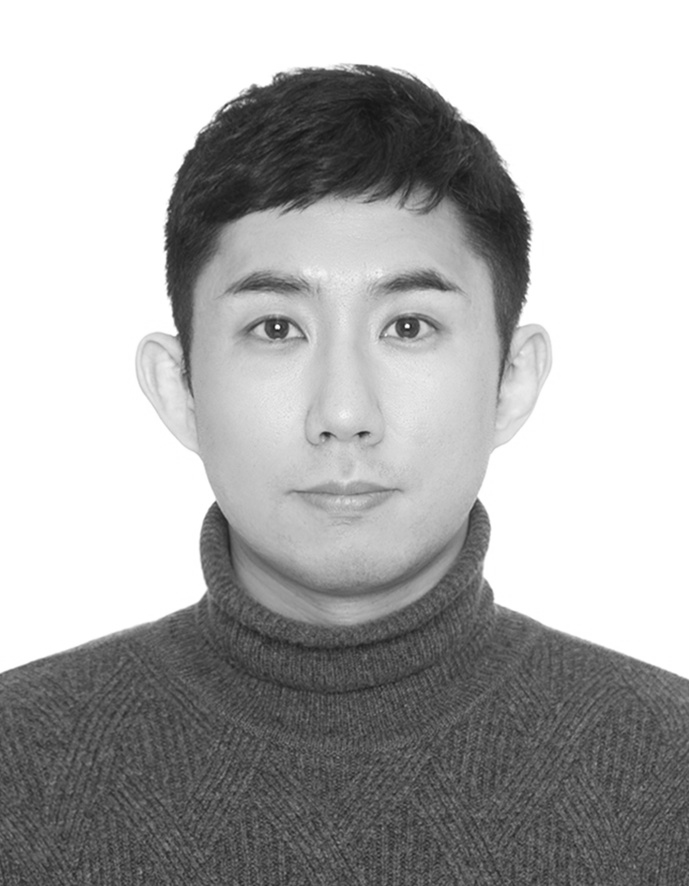}}]{Youngjoon Yoo} received the B.S. degrees in electrical and computer engineering at Seoul National
University, Seoul, Korea, in 2011.
He received a Ph. D. in Electrical Engineering from Seoul National University, Seoul, Korea in 2017.
He is currently a research scientist in NAVER AI Research, and also leads the Image Vision team, NAVER CLOVA. 
His research interests include deep learning for computer vision and probabilistic theory.
\end{IEEEbiography}

\EOD

\end{document}